\documentclass{ieeeaccess}
\usepackage{cite}
\usepackage{amsmath,amssymb,amsfonts}
\usepackage{algorithmic}
\usepackage{graphicx}
\usepackage{textcomp}
\usepackage[numbers]{natbib}
\usepackage{url}

\usepackage{bm}
\makeatletter
\AtBeginDocument{\DeclareMathVersion{bold}
\SetSymbolFont{operators}{bold}{T1}{times}{b}{n}
\SetSymbolFont{NewLetters}{bold}{T1}{times}{b}{it}
\SetMathAlphabet{\mathrm}{bold}{T1}{times}{b}{n}
\SetMathAlphabet{\mathit}{bold}{T1}{times}{b}{it}
\SetMathAlphabet{\mathbf}{bold}{T1}{times}{b}{n}
\SetMathAlphabet{\mathtt}{bold}{OT1}{pcr}{b}{n}
\SetSymbolFont{symbols}{bold}{OMS}{cmsy}{b}{n}
\renewcommand\boldmath{\@nomath\boldmath\mathversion{bold}}}
\makeatother

\def\BibTeX{{\rm B\kern-.05em{\sc i\kern-.025em b}\kern-.08em
    T\kern-.1667em\lower.7ex\hbox{E}\kern-.125emX}}

\begin{document}
\history{Date of publication xxxx 00, 0000, date of current version xxxx 00, 0000.}
\doi{10.1109/ACCESS.2023.1120000}

\title{ExpPoint-MAE: Better interpretability and performance for self-supervised point cloud transformers}
\author{Ioannis Romanelis\authorrefmark{1,2,†},
Vlassis Fotis\authorrefmark{1,†}, Konstantinos Moustakas\authorrefmark{1}   
\IEEEmembership{Member, IEEE},
 and Adrian Munteanu\authorrefmark{2}}

\address[1]{Department of Electrical and Computer Engineering, University of Patras, Greece (e-mails: iroman@ece.upatras.gr, vfotis@ece.upatras.gr, moustakas@ece.upatras.gr)}
\address[2]{Department of Electronics and Informatics, Vrije Universiteit Brussel (e-mail: adrian.munteanu@vub.be, -)}
\address[†]{Indicates equal contribution}
\tfootnote{This work has received funding from the European Union's Horizon 2020 research and innovation program under Grant Agreement No. 101092875 - DIDYMOS-XR: Digital DynaMic and responsible twinS for XR.)}

\markboth
{Author \headeretal: Preparation of Papers for IEEE TRANSACTIONS and JOURNALS}
{Author \headeretal: Preparation of Papers for IEEE TRANSACTIONS and JOURNALS}

\corresp{Corresponding author: Ioannis Romanelis (e-mail: iroman@ece.upatras.gr).}

\begin{abstract}
In this paper we delve into the properties of transformers, attained through self-supervision, in the point cloud domain. Specifically, we evaluate the effectiveness of Masked Autoencoding as a pretraining scheme, and explore Momentum Contrast as an alternative. In our study we investigate the impact of data quantity on the learned features, and uncover similarities in the transformer’s behavior across domains. Through comprehensive visualizations, we observe that the transformer learns to attend to semantically meaningful regions, indicating that pretraining leads to a better understanding of the underlying geometry. Moreover, we examine the finetuning process and its effect on the learned representations. Based on that, we devise an unfreezing strategy which consistently outperforms our baseline without introducing any other modifications to the model or the training pipeline, and achieve state-of-the-art results in the classification task among transformer models.
\end{abstract}

\begin{keywords}
Deep Learning, Explainability, Point Clouds, Self-Supervision, Transformers
\end{keywords}

\titlepgskip=-21pt

\maketitle

\section{Introduction}
\label{sec:introduction}
Deep learning models at large scale require adequately large labeled datasets to be able to learn. This is clearly a limitation for deep learning in general, since manual annotation is a very time-consuming and costly task. As a result, it was a great breakthrough when it was discovered that models can actually benefit from unlabeled data, by using them to design and solve pretext tasks, in which the sample itself is the label. 

This concept appears to have features resembling a real-world analog, that is, the training happening inside a small child’s/infant’s brain. Supervision from experts (adults) is actually responsible for a small percentage of knowledge acquired throughout a human’s lifespan, and it mostly tailors to specific material rather than the perception of the real world at large. Simply by observing the behavior of other people and surrounding objects, they gain intuitive understanding of their environment. While we have fairly successfully replicated a form of supervised learning, self-supervision is an area that still lags behind, despite theoretically offering the most advantages.

Naturally, the data needs to be diverse and of sufficient size for this process to actually yield any benefits. Fortunately, by leveraging the ocean of unlabeled data that is freely available on the web, researchers have managed to significantly boost the performance of their models. Great examples of this concept can be found in both language \citep{BERT} and in image \citep{kriz, He_2016_CVPR} domains, showing great promise for a variety of potential applications in other domains.

The most intuitive and (arguably) popular pretext task applied both in vision and language is the so-called “fill in the gaps”. It refers to tasks where input samples are truncated to occlude pieces of information or corrupted by applying various types of perturbations and noise. The deep learning model is given the damaged sample as input and is tasked to reconstruct/complete the original. During this process, the model learns features related to the sample’s class. These features are incredibly useful and can be leveraged to improve the performance of the model in downstream supervised tasks, involving potentially much smaller datasets.

As this approach became more common, more sophisticated techniques of self-supervision were devised \citep{MoCo, sfclvr, Caron1}. However, few studies have actually delved into what it is that the models actually learn \citep{NEURIPS2021_652cf383, nguyen2021do, Chefer_2021_CVPR} and even fewer have examined point clouds in particular \citep{explainingpointnet, Tayyub_2022_ACCV}.
 Point Cloud models differ significantly from image-based ones. As opposed to an image’s canonical grid structure, uniform density and general data availability, the point clouds are highly irregular, highly inconsistent in terms of density, and are generally scarce compared to images. Additionally, contrary to an image’s fixed pixel position, point clouds are unordered, often different in terms of cardinality, and require that their handling be invariant to permutations. Due to these challenges that arise in the domain shift from language/image to point clouds, we find that this particular area is lacking. 

\color{black}

In this study we adopt the standard transformer as our baseline model. By "standard transformers" we refer to models that follow the architecture presented in ViT \citep{dosovitskiy2020vit}, as opposed to architectures that use transformer-like blocks. It is a highly versatile, strong architecture that has demonstrated incredible results in both language and vision, overtaking previous state-of-the-art models \citep{vitvresnet}. However, transformer-based works on point clouds do not reflect this success. In this work we explore various aspects of this architecture, in order to better understand its inner workings and find ways to improve its performance. We argue that our findings will be of further use in future works, since they are targeted at a widely used, general-purpose architecture rather than a specialized one.

To summarize, our contributions are twofold:

\begin{itemize}
    
    \item We propose \textit{strategic unfreezing}, a finetuning strategy that retains the properties of the backbone, learned through pretraining, while increasing the accuracy of our baseline both in 
    ModelNet40 (+0.5\%) and ScanObjectNN (+0.86\%, +1.73\%, +0.07\%), and achieving state-of-the-art results among transformer models. (Sec. \ref{sec:4})

    \item We adjust explainability tools from the image and NLP domain to work with point clouds, with the aim of understanding the inner workings of the transformer and the effect of pretraining. Interestingly, we uncover that with more data, the transformer seems to learn the inductive bias of convolution, to attend locally.  (Sec. \ref{sec:5})
    
    
\end{itemize}

\section{Related Work}\label{sec2}
\subsection{Deep Learning on Point Clouds}

The domain shift from 2D to 3D came with several extra challenges, due to the lack of grid structure and uneven density of point clouds. Early attempts involved the application of image methods as-is, using multi-view images \citep{multiview1, multiview2} or voxels \citep{pvnet, voxels111}. Not long after that, specialized architectures emerged. PointNet \citep{pointnet} pioneered point-wise MLPs and pooling for extracting global features. PointNet++ \citep{pointnetpp} later followed a multi-scale approach by incorporating neighborhood information. Following this paradigm, other works create more complex kernels, taking advantage of geometric priors \citep{kpconv, rsconv, pointmlp, curvenet, spidercnn}. 

\begin{figure}
\centering
\includegraphics[width=3.26in]{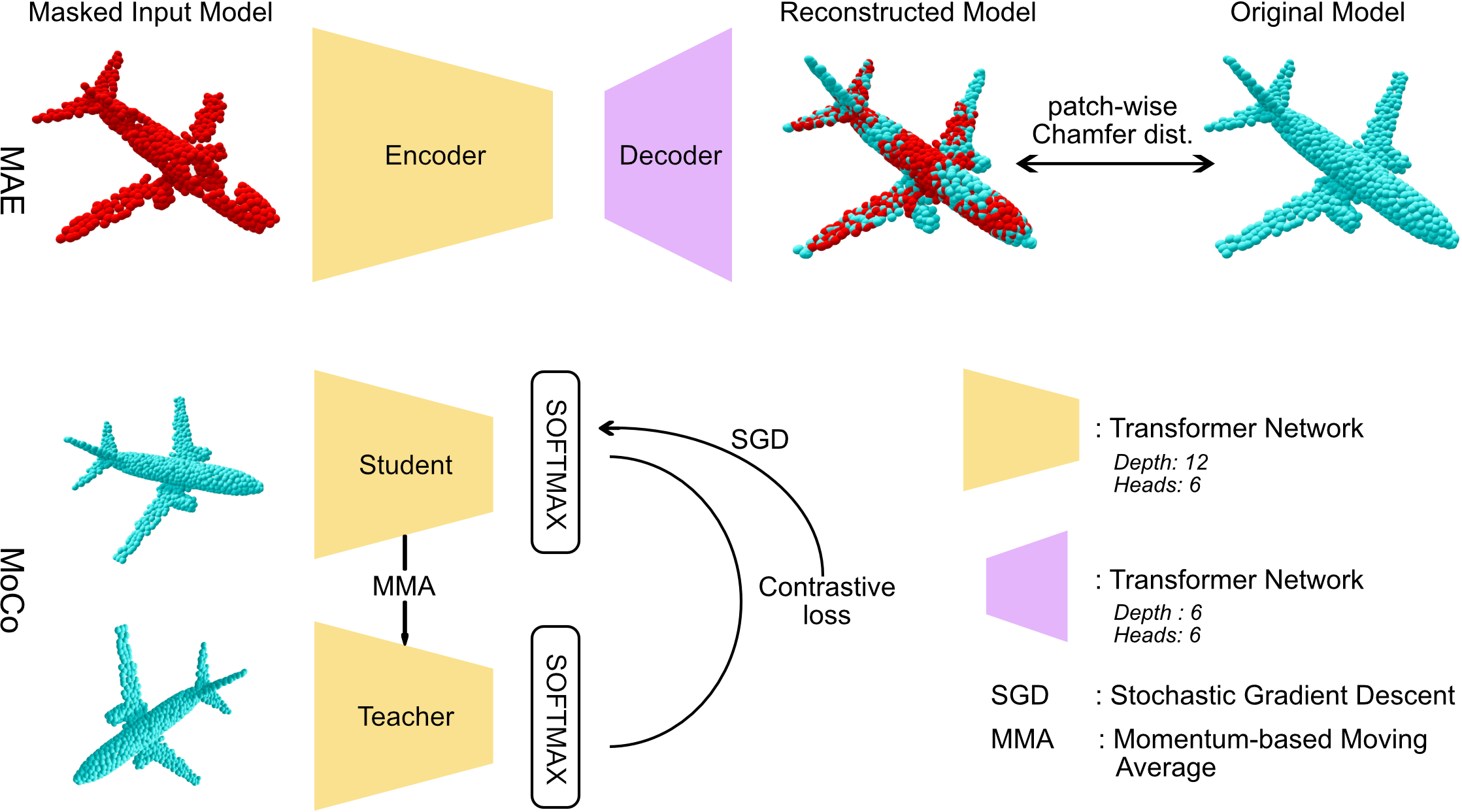}
\caption{Graphical description of the two pretraining pipelines studied in this paper, namely Masked-AutoEncoding (MAE) \citep{mae} and Momentum Contrast (MoCo) \citep{MoCo}. In simple terms, MAE trains an autoencoder to reconstruct a shape with missing parts, whereas MoCo trains two networks (Student/Teacher) to generate approximately equal predictions for different augmentations of a data sample.}
\label{fig:reconstruction}
\end{figure}

More recently, transformer variants have been successfully applied to the point cloud domain. Originally introduced in  \citep{attention_is_all_you_need}, transformers were quickly established as the go-to choice in NLP. Dosovitskiy et al.\citep{dosovitskiy2020vit}, later introduced the vision transformer, operating on image data. This was accomplished by tokenizing the input into patches and incorporating spatial information through a positional embedding. This approach also allowed data from different modalities to be used jointly \citep{Multimodal, Multimodal2}, making the transformer the most prominent architecture in that field. Point Transformer \citep{Zhao_2021_ICCV} is among the first works to apply a transformer-like architecture to point cloud data. They apply intra-neighborhood attention to create a patch feature vector and use pooling operations to downsample the point cloud.

In PCT \citep{Guo_2021}, they extract point embeddings through a downsampling network and apply a series of single attention transformer layers to the remaining points. 
The feature maps of these layers are concatenated, and global features are extracted through mean and max pooling. In PVT \citep{PVT}, the authors create specialized attention modules for both points and voxels, utilizing the complementary information they contain to extract better features.

Point-BERT \citep{Yu_2022_CVPR} tries to generalize the plain transformer, as used both in Image and NLP. They train a 'tokenizer' autoencoding network, based on DGCNN \citep{dgcnn}, to map point neighborhoods into feature vectors. During pretraining, a transformer encoder-decoder architecture receives masked point clouds as input and tries to reconstruct the embeddings of the 'tokenizer'. Additionally, the classification token is trained through a contrastive loss like in \citep{MoCo}. 
Point-MAE \citep{mae} and Point-Multiscale MAE (M2AE) \citep{He_2022_CVPR} both utilize masked autoencoding to pretrain their transformer backbones, by reconstructing the actual points of the masked neighborhoods directly. Point-M2AE uses a pyramid-like backbone that gradually downsamples the input point cloud, obtaining multi-scale features. On the other hand, Point-MAE uses the same architecture as Point-BERT, that is, the plain transformer. 

While hierarchical architectures such as M2AE are typically associated with slightly better performance, they require parameter tuning when transferred to other domains or scaled to larger datasets. Since the main scope of this work is explainability, the simplicity and cross-domain applicability of MAE are more valuable. It can be trivially scaled by adding or removing blocks and provides a fair ground for comparisons with image and NLP. Therefore, for our baseline model, we choose to adopt the transformer presented in Point-MAE \citep{mae}. 


\subsection{Self supervised Learning} 
The idea of making use of large amounts of unlabeled data to improve a learning model is not new, rather it has been around for decades \citep{lecun1, ng, transfer, denoising}. Its popularity did not surge until the mid-2010s \citep{lecun2, kriz, He_2016_CVPR} however, when the availability of unlabeled data at scale started to increase and the potency of hardware could finally keep up to the challenge. Of particular importance is masked autoencoding, pioneered by \citep{BERT}, which revolutionized language modeling. This paradigm was later followed by other works, such as \citep{fewshot, gpt2, zhai2022lit}, that further scaled up  in terms of parameter and dataset size, achieving state-of-the-art results in the few-shot and zero-shot settings. In \citep{fractaldb} they experiment with pretraining by using synthetic data, artificially injected with desirable properties.  

Another line of work that has been amassing popularity recently is contrastive learning \citep{MoCo, sfclvr, mocov3, mocov2, dino, Jiang_2021_ICCV, rlcpc} as a way to learn meaningful data representations. It is accomplished by generating positive and negative sample pairs and training the model to pull together or push them away, respectively. The pairs are generated through data augmentations; perturbations, crops, and transformations. A similar idea, dubbed contrastive clustering, is applied to groups of samples instead of pairs \citep{Caron1, Caron_2018_ECCV}. \citep{Caron1} in particular, eases up the computational load by introducing learnable cluster centers. In \citep{BYOL}, the authors eliminate the need for negative pairs by training two networks to produce matching feature vectors for two different views of the same sample.

Despite the scarcity of point cloud data compared to image and language data, and the corresponding need for good self-supervision techniques in this domain, the topic has only recently begun attracting attention. \citep{Eckart_2021_CVPR} proposes to split point clouds into parts and use these parts to parameterize gaussian mixture models. The model is then trained by using a loss that resembles likelihood maximization. In \citep{HNS} the authors perform patch-level contrastive learning. By rotating a query patch they form positive pairs, while negative pairs are formed by taking into account any other patch of the shape. A similar approach is followed in \citep{zeroshot}, finding positive and negative pairs by using an inductive model to perform pseudolabeling. In \citep{Wang_2021_ICCV}, a similar approach to masked autoencoding is followed, that is, completion of occluded point clouds acquired by taking custom viewpoints through virtual cameras. \citep{pointcontrast} argues that performing pretext tasks on single object point clouds might have limited benefits in real world applications. They instead sample positive and negative pairs from complex scene scans, hoping to get a better estimate of the target distributions, and apply a contrastive loss to train their model.

\subsection{Explainability in deep learning}

An extremely important field in deep learning research is explainability. Deep models are generally seen as black boxes and any attempt at designing new or improving existing models is usually empirical. Nevertheless, there are several tools one can use in order to better understand a model's behavior. In \citep{CKA1, CKA2} the authors present CKA, a similarity measure between feature representations of two arbitrary neural networks. \citep{NEURIPS2021_652cf383} utilizes this tool and provides valuable insights into the differences between how CNNs and transformers learn, while \citep{wide} tries to compare the representations of networks with different depths and feature map sizes. Other works \citep{Tayyub_2022_ACCV, Chefer_2021_CVPR} utilize gradient-based methods in order to visualize the receptive fields or the relevancy of input patches towards the model's decision. 
We use a combination of all the above tools to attain a complete picture of our model and accompanying pretraining scheme.

\section{Baseline Method, Tools, and Dataset}
 
In this section, we briefly overview our baseline model and pretraining setup. After training and finetuning, we assess the quality of the learned features both qualitatively and quantitatively by using explainability tools and measuring the accuracy in the classification task. Further details regarding the model and the training process can be found in the appendix. We utilize pytorch lightning\footnote{\url{https://www.pytorchlightning.ai/}} for our codebase, in favor of clarity, transparency, and reproducibility. The code is publicly available on github\footnote{\url{https://github.com/VVRPanda/ExpPoint-MAE.git}}.

\subsection{Masked AutoEncoders}
For our baseline model, we adopt the vision transformer trained with masked autoencoding, as presented by Pang et al. \citep{mae}, which we restate here for the sake of completeness. The transformer is comprised of 12 blocks with 6 attention heads per block.
The input point cloud is split into N, possibly overlapping patches, 60\% of which are masked out. The remaining patches are embedded into feature space via a small PointNet-like \citep{pointnet} network, and the patch centroids are used to generate positional encodings. The two are added to form the input to our transformer, which outputs a feature vector for each patch. At this stage, masked patches are assigned a mask token and a positional encoding and are concatenated with the rest before being fed into the decoder. The decoder reconstructs the masked patches and a reconstruction loss based on Chamfer distance is applied patch-wise. 

\begin{figure}
    \centering
    \includegraphics[width=3.25in]{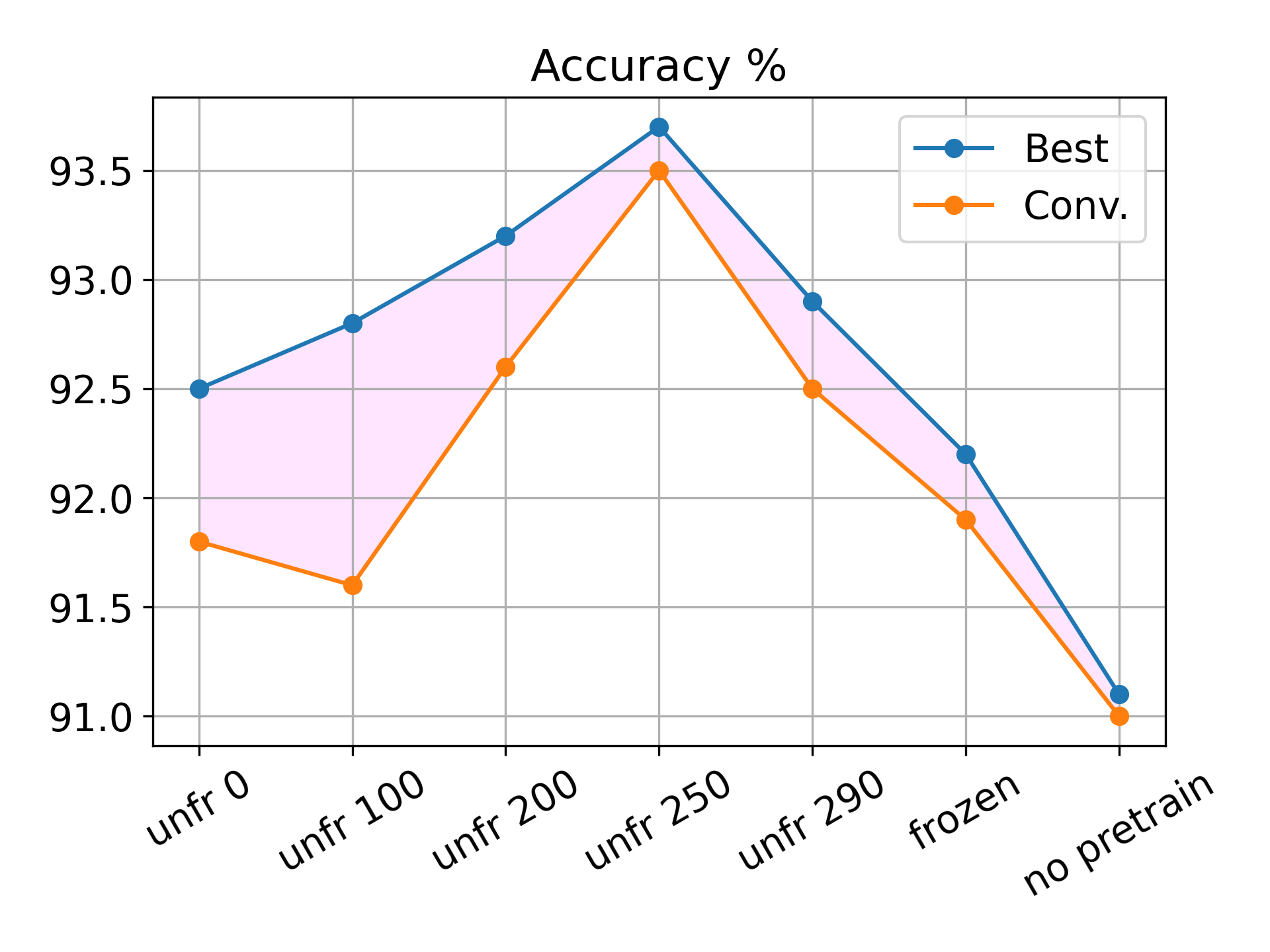}
    \vspace{-0.2cm}
    \caption{Comparison of different unfreezing points for the backbone of the transformer model. Unfreezing the model too early or too late can result in suboptimal results, as the network may either 'forget' the features learned through pretraining or fail to acquire task-specific knowledge, respectively.}
    \label{fig:unfreezing}
\end{figure}

\subsection{Explainability Tools}
For explainability, one of the main points of interest in this work, we utilize a variety of tools. The Centered Kernel Alignment criterion offers a versatile way of comparing feature representations between two models across a dataset, and its importance to our work cannot be overstated. It provides insights into the differences between models that have been pretrained \& finetuned vs just finetuned, as well as models that have been pretrained using varying datasets and or strategies. We omit the math behind CKA and refer the interested reader to \citep{CKA1, CKA2}. Attention visualization is another commonly used tool that helps to understand how the model correlates the point patches based on the extracted geometric features. Finally, we compute and visualize the receptive fields to better understand the information flow within the network.

\subsection{Dataset}

As opposed to their image counterparts, point cloud datasets are generally fewer and significantly smaller in size. Additionally, there are a lot of impactful differences between point cloud datasets, including sampling density, scanning device-specific artifacts, and shape variations in general (real-world objects, CAD models). As a result, a model trained on a specific dataset may not generalize to other data distributions. Our first goal towards explainability is to see how the amount of training data might affect the network's performance. To this end, we concatenate the commonly used ShapeNet \citep{shapenet} with the more obscure CC3D \citep{Cherenkova_2020} (41k and 43k training samples, respectively). CC3D contains high resolution, single-object, fine-grained CAD shapes from arbitrary categories. 
To study the effects of the increased volume of data, we conduct pretraining experiments using only ShapeNet (\textbf{S}) and the concatenated dataset (\textbf{C+S}). We compare the two by finetuning for classification in ModelNet40 \citep{modelnet40paper} and ScanObjectNN \citep{scanobjectnnpaper} and report the results in Tables \ref{t:ModelNet40}, \ref{t:ScanObjectNN}. In both cases, our proposed dataset helps achieve better accuracy. For the rest of the paper, we will be using \textbf{C+S} for pretraining, unless otherwise specified.

\section{Strategic Unfreezing}
\label{sec:4}

\begin{figure*}[!t]
\centering
\includegraphics[width=6.7in]{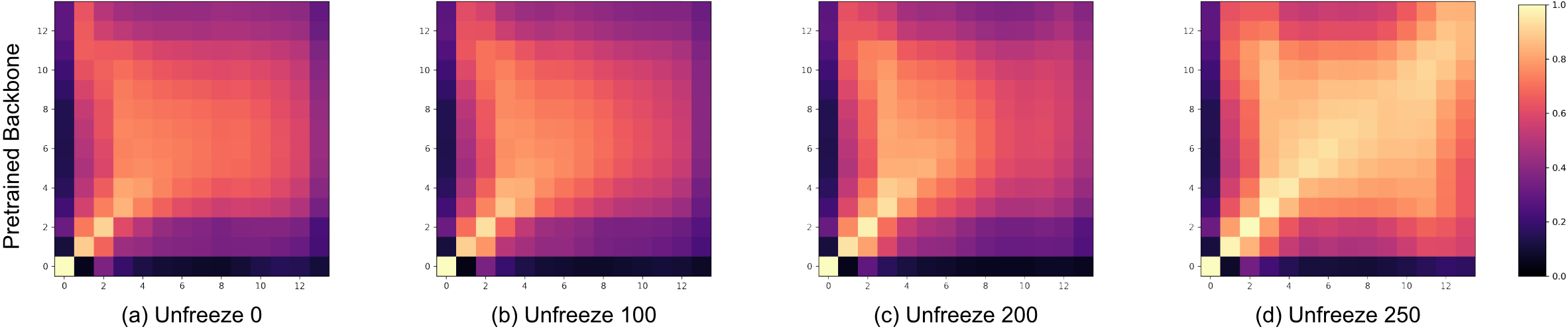}
\caption{CKA comparison of the pretrained backbone (y-axis) with  versions that have been finetuned for 300 epochs, 
  unfreezing the backbone on various epochs (x-axis). The first and second blocks indicate the positional and feature embedding extractors, while the rest of the blocks correspond to the outputs of the attention layers. High values indicate high similarity between feature representations. The later  the network is unfrozen, the higher the similarity with the pretrained backbone, retaining the properties learned through pretraining. In the case of (a), as done in \citep{mae}, the final network has very little similarity with the pretrained backbone, thereby nullifying the effectiveness of pretraining.}
\label{fig:unfreezeCKA}

\end{figure*}

\noindent There are several ways to approach the finetuning process of a pretrained model. Most commonly, the pretrained backbone is frozen, that is, gradient propagation is allowed but no weight updates happen, and a smaller classification (or other) head is trained to map the learned features to logits.
This method is based on the assumption that the backbone has already learned a robust representation of the data and is capable of separating the class clusters effectively. However, the learned data representation often differs significantly from the data used in the downstream tasks. A great example of this is pretraining on synthetic data and using the model in real-world applications. 

Another approach is to perform task-specific adjustments to the model by training both the backbone and the added head end-to-end using a small learning rate. However this contains an important caveat, as the random initialization of the classification head's weights can cause the weights of the backbone to be perturbed in an unintended direction, effectively destroying the learned features and representations. 

In order to properly evaluate these approaches we perform comparative the following experiments.

\begin{itemize}
\item We test the first method by measuring the accuracies in ModelNet40 classification (Figure \ref{fig:unfreezing}). We see that when keeping the backbone frozen and only relying on the classification head, the accuracy is vastly inferior compared to the backbone being unfrozen at various intervals. This is because no task-specific knowledge has been incorporated into the backbone.

\item In Figure \ref{fig:unfreezeCKA}(a), we test the second method. We compare the pretrained backbone that has been finetuned without freezing and the pretrained backbone without finetuning. It is evident that there is hardly any similarity between them, except for the very early layers. This arguably defeats the purpose of pretraining, since very few of the valuable properties are retained in the final network. This deficiency is also reflected in the final accuracy score, as seen in Figure \ref{fig:unfreezing}.

\item Finally, in Figure \ref{fig:unfreezing} we note the accuracy of the model that has not been pretrained. It exhibits the lowest accuracy among all models that have been pretrained, proving the effectiveness of pretraining.

\end{itemize}


Before diving into what the model actually learns (section 6), we first propose a two-stage training scheme that retains the pretrained network's characteristics and obtains superior performance. First, we train the classification head for the majority of epochs, ensuring that it has learned to separate the clusters formed via the backbone. Finally, we jointly train both of them for the remainder of the epochs. In order to figure out the appropriate step to unfreeze the backbone, we perform several experiments and compare the feature representations with the pretrained network each time (Figure \ref{fig:unfreezeCKA}).  Additionally, we measure the best and the convergence accuracy for each of these models in the classification task on ModelNet40 (Figure \ref{fig:unfreezing}). It is evident that the earlier the backbone is unfrozen, the more the pretrained backbone's features are distorted, and the final accuracy is compromised.

In order to identify the proper unfreezing epoch, one must consider the data and the task. For ModelNet40 classification, as a subset of ShapeNet, similar samples were seen by the model during pretraining. As a result, unfreezing during the late stages is appropriate. On the other hand, when finetuning on ScanObjectNN, that consists of real scanned objects, it is wiser to unfreeze earlier so that the backbone can become accustomed to the new data distribution. Finally, when finetuning on the hardest variant of the ScanObjectNN which includes other modifications, such as rotations, we found that the most effective way is to perform an extra domain adapaptation pretraining step, before finetuning. The results of this step can be seen in Table \ref{t:ScanObjectNN} in the SoNN entry. More details are presented in the Appendix. 

Designing and training a model is a difficult and iterative process, that is based heavily on parameter tuning. We believe that with our method researchers will be able to make more informed decisions that will allow them to utilize their models to their maximum capacity. 

\renewcommand{\arraystretch}{1.2}
\begin{table}[h]
\caption{Comparison of different transformer based models and pretraining schemes with and without using voting on the ModelNet40 test set.}
\label{t:ModelNet40}
\begin{tabular}{l | c c}
MODEL  &  W/O VOTING & VOTING\\ 
\hline
 Point-BERT \citep{Yu_2022_CVPR}  & 92.7 & 93.2\\ 
 MaskPoint \citep{liu2022masked} & - & 93.8 \\
 Point-MAE \citep{mae}  & 93.2 & 93.8 \\ 
 PCT-2L \citep{Guo_2021}  & 93.2 & -\\
 PCT-3L \citep{Guo_2021} & 93.4 & -\\
 Point-M2AE \citep{zhang2022point}  & 93.4 & \textcolor{cyan}{94.0}\\
 Ours (\textbf{S})  & \textcolor{cyan}{93.6} & -\\
 Ours (\textbf{C+S}) & \textcolor{blue}{93.7} & \textcolor{blue}{94.2}\\ 
\hline 
\multicolumn{3}{p{170pt}}{\textcolor{blue}{Blue} and \textcolor{cyan}{Cyan} denote 1st and 2nd highest.} \\
\end{tabular}
\end{table}

\begin{table}
\caption{Classification results on ScanObjectNN. The HARD variant refers to PB-T50-RS that is the most challenging one.}
\label{t:ScanObjectNN}
\begin{tabular}{l|c c c}
MODEL & OBJ-BG & OBJ-ONLY & HARD  \\ 
\hline
 Point-BERT & 87.43 & 88.12 & 83.07 \\
 MaskPoint & 89.3 & 89.7 & 84.6\\
 Point-MAE & 90.02 & 88.29 & 85.18 \\
 Point-M2AE & \textcolor{blue}{91.22} & 88.81 & \textcolor{blue}{86.43} \\ 
 Ours (unfr 200) & \textcolor{cyan}{90.88}  & 89.33 & 84.98\\ 
 Ours (unfr 250) & 87.44 & \textcolor{blue}{90.02} & 83.80\\ 
 Ours (SoNN\footnotemark[1]) & 90.36 &  \textcolor{cyan}{89.67} & \textcolor{cyan}{85.25} \\
\hline 
\multicolumn{4}{p{210pt}}{\textcolor{blue}{Blue} and \textcolor{cyan}{Cyan} denote 1st and 2nd highest.} \\
\multicolumn{4}{p{210pt}}{\footnotemark[1]: SoNN indicates that the model has been pretrained for a few epochs in the ScanObjectNN-hard dataset.} \\
\end{tabular}
\end{table}

\section{Explainability Study}
\label{sec:5}

\begin{figure*}
    \centering
    \includegraphics[width=6.7in]{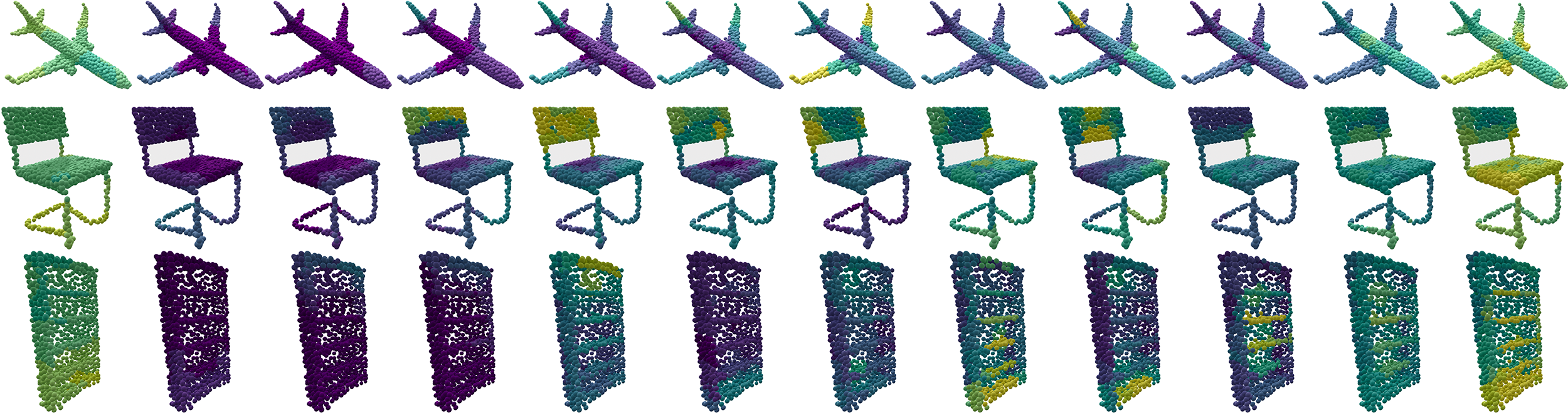}
    \caption{Attention Visualization of the classification token for each block (1-12), averaged across heads (brighter = higher score). Although the classification token's attention score towards itself cannot be visualized, it has the highest value in all cases. This score is included in the normalization process, so that the relative scale between them is visible.}
    \label{fig:att_vis}
\end{figure*}

\begin{figure*}
    \centering
    \includegraphics[width=6.7in]{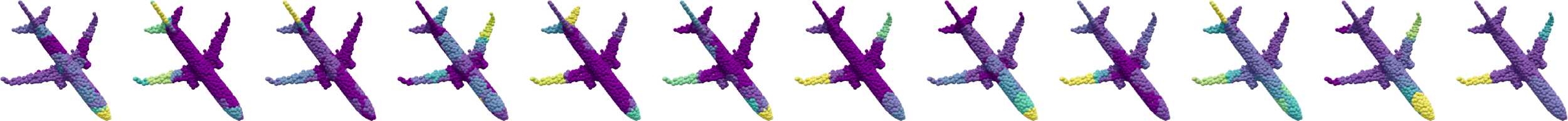}
    \caption{Attention Visualization of the classification token for each block (1-12), averaged across heads for a finetuned model \textbf{without} pretraining. It is evident the locations it attends to do not follow any recognizable pattern. In each layer, the attention is focused on a specific location, which is often the same between layers (e.g. head in layers 0,5,7,10).}
    \label{fig:att_vis_np}
\end{figure*}

\subsection{Attention Visualization}

In this section, we will use the finetuned backbone, pretrained with MAE, and finetuned with the strategy outlined in the previous section. This is because it retains the useful properties acquired through pretraining, but also has a trained classification token, as opposed to \citep{mae}, which uses max-average pooling. By visualizing the attention matrices at each block we can obtain meaningful information regarding how the model correlates various parts of the shape. This information can help identify patterns that indicate whether the pretraining procedure was successful, such as semantic correlation, symmetry, and locality.

We observe that in all input shapes, the first attention block is always global, while the rest gradually shift from attending to specific regions to attending globally. Sharp geometric features in particular attract high attention scores in most layers (Figure \ref{fig:att_vis}). This behavior is quantitatively verified through attention distances in the following section. 

It is important to note that finetuned models that have not been pretrained \textbf{do not} share this trait. As can be seen in figure \ref{fig:att_vis_np}, although the classification token attends to locations with
valid semantic meaning, the behavior is erradic, repetitive and there is no clear transition from specific areas to the whole shape.

\subsection{Attention Distance}
Following the paradigm of \citep{NEURIPS2021_652cf383}, we study the sorted attention distance. The mean Attention Distance (mAD) of layer $l$ for a head $h$ is calculated as follows: 

\begin{equation*}
    mAD_{l,h} = \frac{1}{N^2} \sum_{ij} \mathbf{A_l} \odot \mathbf{D}
\end{equation*}

Where $\mathbf{A_l}^{N \times N}$ is the attention matrix of layer $i$ and $\mathbf{D}^{N \times N}$ is the distance matrix, where each entry $d_{i,j}$ is the Euclidean distance between centers $i$ and $j$ and the $\odot$ symbol denotes the Hadamard product. The product is averaged across all entries to obtain a mean attention-distance value, which is also averaged across the validation set of ModelNet40.

\begin{figure*}[!t]
    \centering
    \includegraphics[width=6.7in]{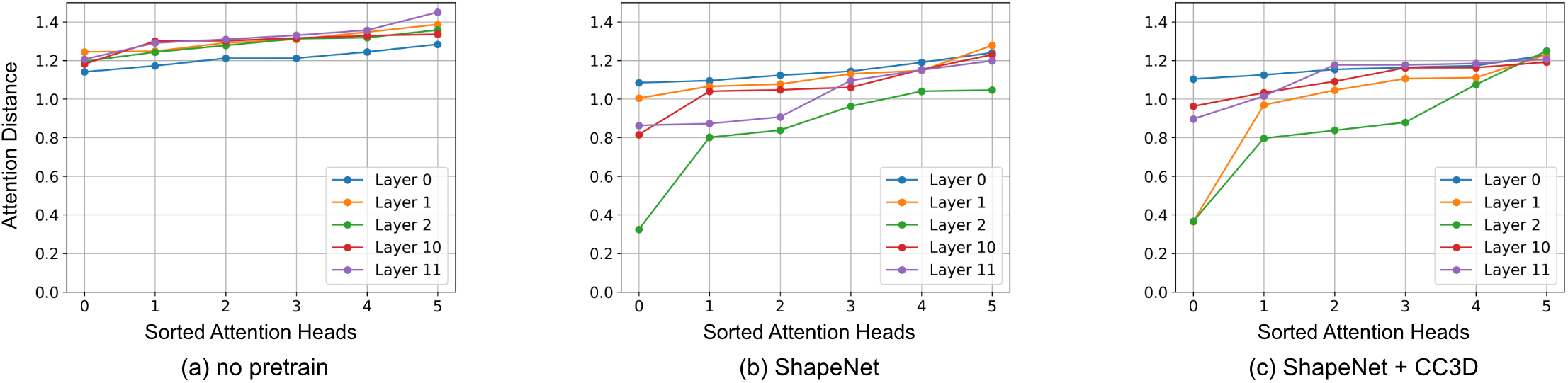}
    \caption{
Attention Distance of the attention heads 
(distances are sorted in ascending order for clarity). The distances are averaged across the entire validation set of ModelNet40. 
We see that as the amount of data increases, early layers of the network begins to attend locally, while higher layers incorporate global information, exhibiting similar behaviour to ViT in the image domain\citep{dosovitskiy2020vit} 
Notably, a key difference is observed in the very first layer, where all heads consistently attend globally.}
\label{fig:attention_distance}
\end{figure*}

Based on the graphs in Figure \ref{fig:attention_distance}, it can be inferred that with the increase in the amount of training data, the earlier layers 
tend to focus more locally with certain attention heads, while others still maintain a global focus.
This observation suggests that the transformer network may begin to learn the inductive bias of the convolution, that is, attending to local features, while still retaining the capability to aggregate global information, thereby resulting in more comprehensive representations. This mitigates the general struggle of self-attention to model local relationships \citep{LSAlimits}, without requiring any modifications to the architecture.

Interestingly, our findings are consistent with those of \citep{NEURIPS2021_652cf383, steiner2021augreg, VITAEV2}, indicating that transformers may exhibit akin behaviors across domains. A noticeable difference is that in the case of ShapeNet (Figure \ref{fig:attention_distance}b) the first two layers seem attend globally with all of their attention heads, whereas when CC3D is added to the pretraining data, this phenomenon is only observed in the first layer (Figure \ref{fig:attention_distance}c). We assume that the network requires a rough understanding of the entire shape before opting to extract local features. This demeanor should be studied further if point cloud datasets at large scale become available in the future.

\subsection{Effective Receptive Fields}

\begin{figure}
    \centering
    \includegraphics[width=3.25in]{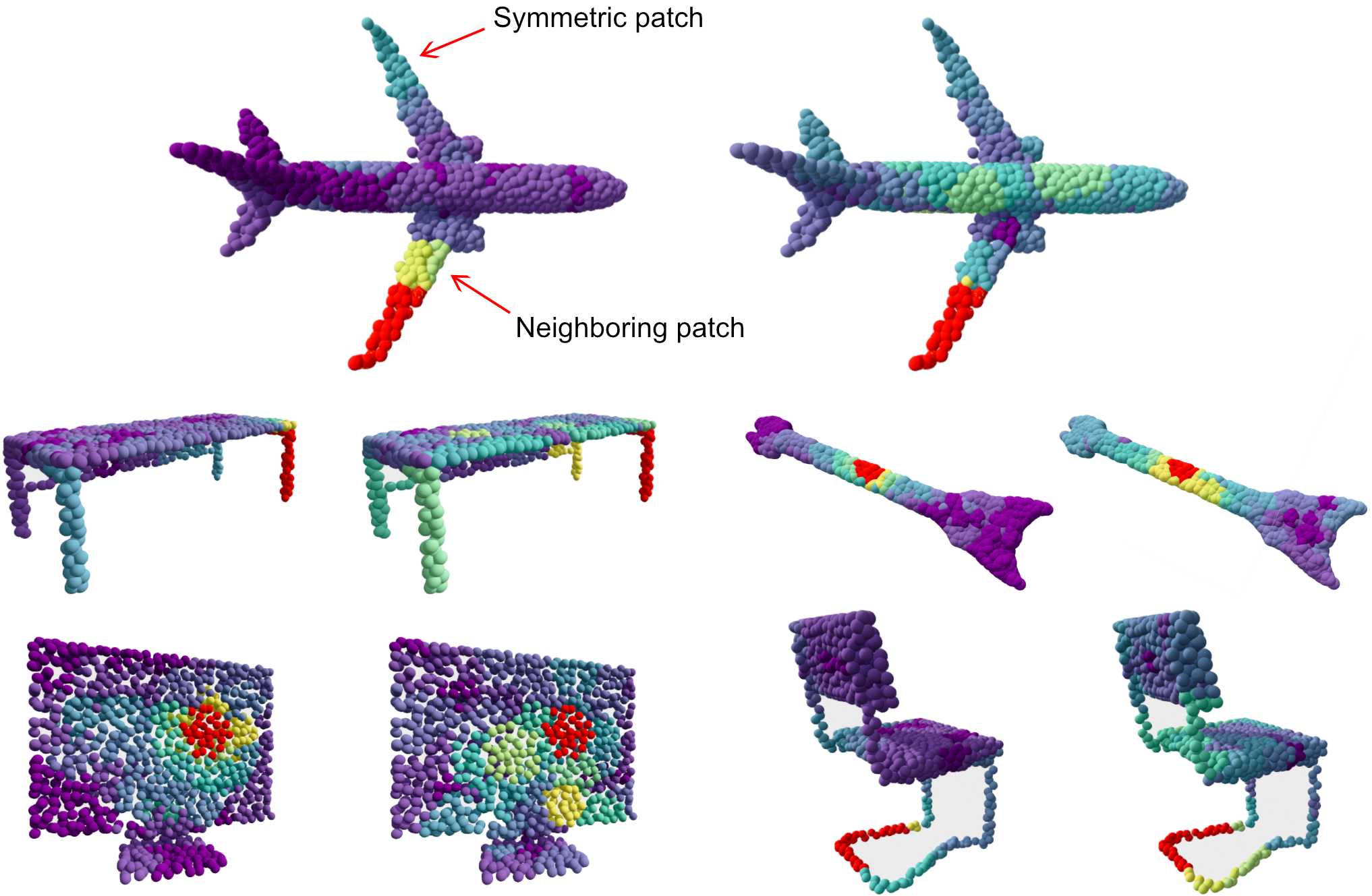}
    \caption{Effective receptive fields of the patch of interest (colored in red). We compute the gradient of the feature representations generated by the last layer with respect to the input patches. Each point is colored based on the magnitude of the gradient of its corresponding patch.
    In the right variant, that is finetuned for classification, the model  retrieves infrormation from the entire shape, as it tries to incorporate class information. Conversly, the left variant, that is the pretrained backbone, focuses mostly to local and symmetric parts, aligning with the task it was trained for, that is reconstruction.}
    \label{fig:receptive_fields}
\end{figure}

We have established that by using our unfreezing strategy, the final network shares many similarities with the pretrained network. In order to understand their differences (trained for pretext versus downstream tasks), we visualize the effective receptive fields \citep{NEURIPS2021_652cf383}. We accomplish this by selecting a patch and propagating the gradients backwards from its feature vector to its embedding, including both feature and positional information. We then visualize the norm of said gradients and present the results in figure \ref{fig:receptive_fields}. 

We notice that the pretrained model, trained explicitly for reconstruction, looks strictly at neighboring as well as semantically similar patches, since they contain the most relevant information. These areas have the highest impact on the finetuned model as well, but its receptive field is noticeably wider. We believe this is because the classification task requires global features.

\section{Contrastive Learning}
\label{sec:6}

Having delved into the inner workings of masked autoencoding and the properties associated with it as a pretraining scheme, we need to evaluate its effectiveness against other pretraining methods in the literature. We choose what we believe to be the next most popular one,  \textit{contrastive learning}. Impressed by its performance in the image domain, we follow the approach of \textit{Momentum Contrast  (MoCo) }\citep{MoCo}, which we extend to work with 3D data. 

In MoCo, two copies of the network are trained simultaneously, one through backpropagation and the other using a momentum update rule. Positive pairs are created by feeding crops of the same shape with different augmentations to these networks, while previous activations of the momentum network act as negative pairs. We follow the same strategy and create our crops by using a mix of gaussian noise, anisotropic scaling, rotations, and random point dropouts. 

By finetuning the models pretrained with MoCo, we observe that, although it improves the original model without pretraining, there is a significant gap in accuracy compared to its MAE counterpart (Table \ref{tab:MAE_MoCo_acc}). This, however, does not rule out the possibility that other variants of contrastive learning might be more effective. Instead, it suggests that a well-designed pretraining scheme will most likely offer a significant performance boost to any baseline model.  

\begin{figure}
    \centering
    \includegraphics[width=3.25in]{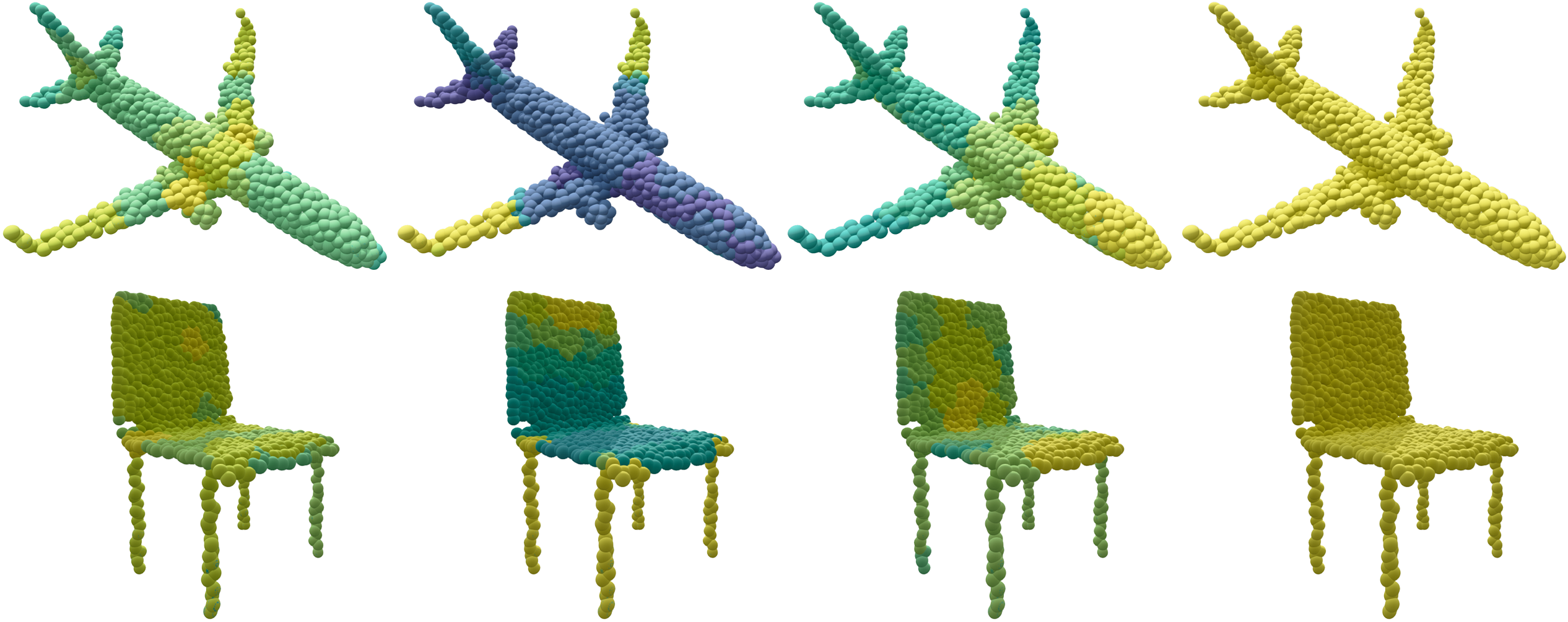}
    \caption{Attention scores of the classification token, averaged across all heads, pretrained using the \textbf{MoCo} pipeline. We visualize a subset of the network layers, [1, 3, 7, 8] left to right. While in the early layers the CLS token attends to meaningful parts of the shape, exhibiting a tendency to attend at symmetric parts, from layer 8 and onwards it attends uniformly to the whole shape.}
    \label{fig:moco_attention}
\end{figure}

\begin{figure}
    \centering
    \includegraphics[width=3.25in]{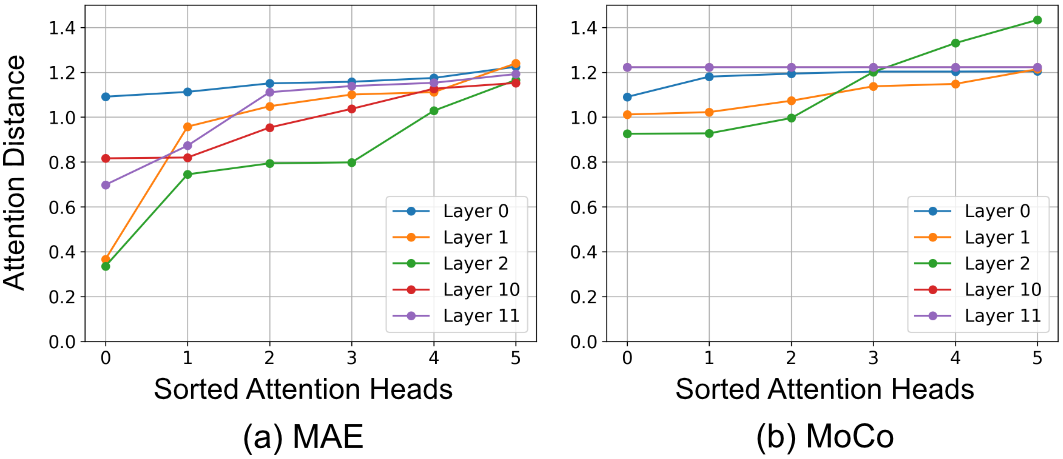}
    \caption{Comparison of Attention Distances achieved through Masked AutoEncoding (MAE) \citep{mae} and Momentum Contrast (MoCo) \citep{MoCo} pretraining schemes. We observe that the setup of MoCo is insufficient to enable the transformer to learn to attend locally. Layers 10 and 11 in MoCo have the exact same attention distance. }
  \label{fig:attention_distanceMOCO}
\end{figure}


\begin{figure*}
    \centering
    \includegraphics[width=6.7in]{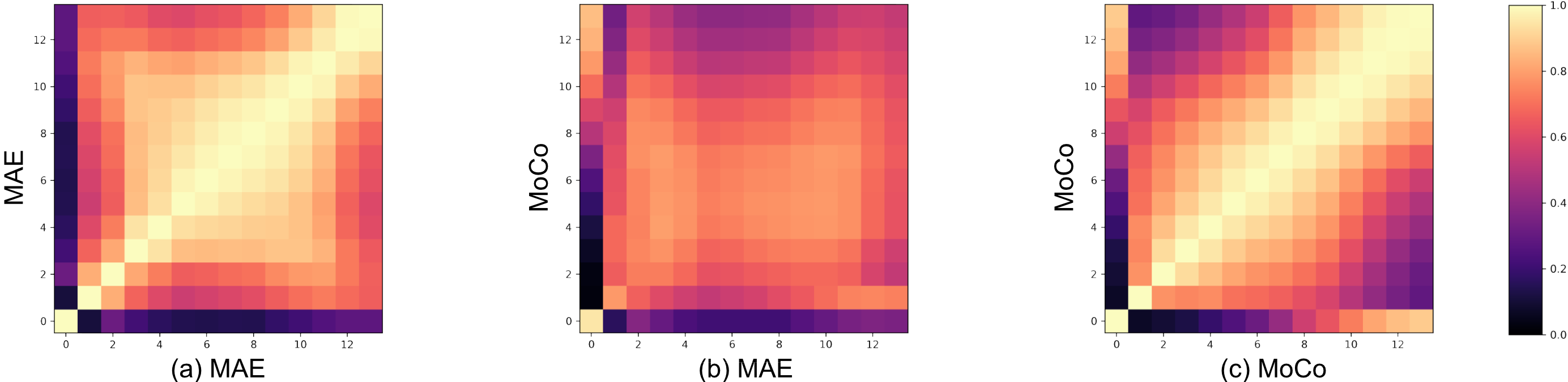}
    \caption{CKA comparisons between the two pretrained backbones (MAE and MoCo pipelines) and themselves. The first two entries correspond to the positional and feature embedding respectively while the rest to the outputs of the attention blocks. We observe little resemblance between the representations learned through the different pretraining pipelines. Interestingly, the final feature representations obtained through the MAE pipeline appear to rely more on the features learned within the layers of the network, showing little similarity with the positional embeddings. Conversely, the MoCo pipeline demonstrates a higher reliance on the positional embedding. }
\label{fig:maemocoCKA}
\end{figure*}


\begin{table}[h]
    \centering
    \caption{ Network Accuracy Achieved with different pretraining schemes
    }
    \label{tab:MAE_MoCo_acc}
    \begin{tabular}{c|c c c c c}
        Model    &  No Pre. & MAE            & MoCo & MAE \& MoCo \\
        \hline
        Accuracy &  91.1    & \textcolor{blue}{93.7}  & \textcolor{cyan}{92.2} & 91.8
    \end{tabular}
\end{table}

As opposed to masked autoencoding, an interesting property of contrastive learning is that it allows the classification token to be trained during the pretraining stage. In fact, by examining the attention scores, we notice that the classification token attends to characteristic areas of the input shape, and exhibits symmetry, by attending to shape parts with similar semantics (Figure \ref{fig:moco_attention}). We also observe that from layers 8 and onward, the CLS token attends uniformly to the whole shape. This is verified by Figure \ref{fig:attention_distanceMOCO}b where layers 10 and 11 have the exact same attention distance across all heads. Finally, we see that through MoCo pretraining the backbone does \textit{not} learn to attend to neighboring patches, indicating that the training scheme might require more data or alterations to converge. For the sake of completeness, we also tested a MoCo and MAE hybrid (details in the Appendix). 

Furthermore, useful insights can be extracted by comparing the two pretraining schemes against each other, as well as themselves, to get an idea of how the information is distributed throughout the layers. The results can be seen in Figure \ref{fig:maemocoCKA}.

We observe that middle layers in MAE seem to have similar feature representations, while there is some similarity between input and output layers. However, the same does not seem to apply to MoCo, where each layer has common representations only with a few of its preceding and succeeding layers, creating a chain-like representation. Astoundingly, the layers seem to gradually lose their similarity with the early feature embedding and rely more and more on the positional embedding, whereas in MAE, the feature embedding is the most dominant in all layers. Naturally, MAE and Moco are based on fundamentally different principles, so they bear very little resemblance and only in the middle layers.

\section{Ablation Studies}
In the ablation study section we focus on two separate topics that are the masking ratio and the intermediate representations of the transformer layers. 

 Firstly, we focus of the reconstruction abilities of the backbone, trained through the MAE pipeline. Point-MAE has already studied the effect of masking in the final accuracy, concluding that a 60\% masking is the optimal choice. In our experiments we want to test the reconstruction ability of the network when masking the input with a higher ratio. As depicted in Figure \ref{fig:reconstruction}, even in the extreme case when 90\% of the model is masked out, the network can provide a decent reconstruction. We believe that this is because positional embeddings of the masked patches are given to the decoder, providing useful hints regarding the position of individual parts.

\begin{figure}[!t]
  \begin{center}
   \includegraphics[width=3.15in]{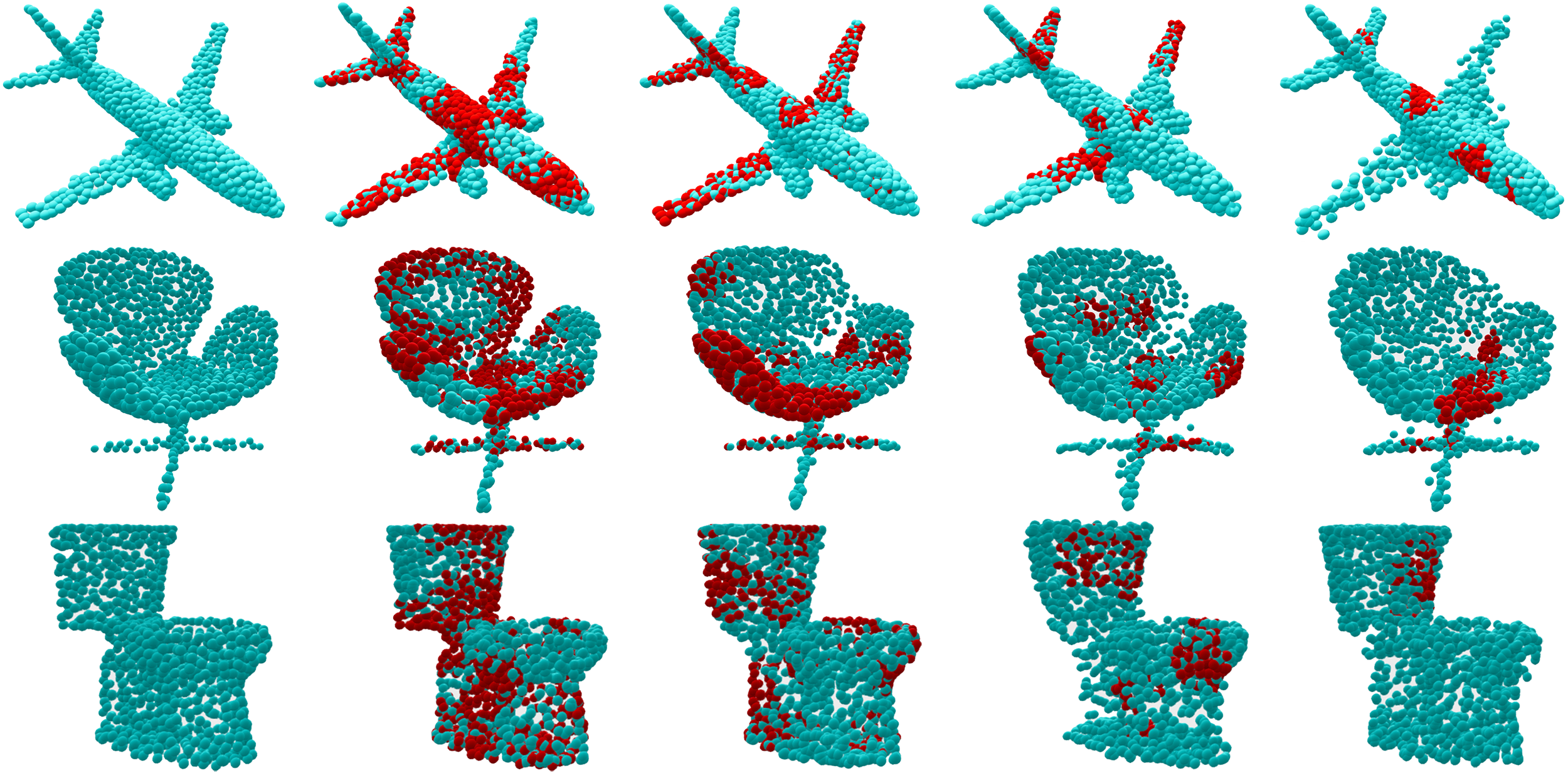}
   \end{center}
    
    \hspace{0.1cm} Original \hspace{0.5cm} 60\% \hspace{0.5cm} 80\%  \hspace{0.6cm} 90\% \hspace{0.5cm} 95\% \\
 \centering
  \vspace{-0.2cm}
  \caption{Reconstruction results using different masking ratios. The network is pretrained, using MAE, with a 60\% masking ratio. We test the reconstruction ability of the system when we provide as input a point cloud with higher masking ratios.  Input and reconstructed patches are shown in red and cyan, respectively.}
  \label{fig:reconstruction}
\end{figure}

\begin{figure}[!t]
    \centering
    \includegraphics[width=3.25in]{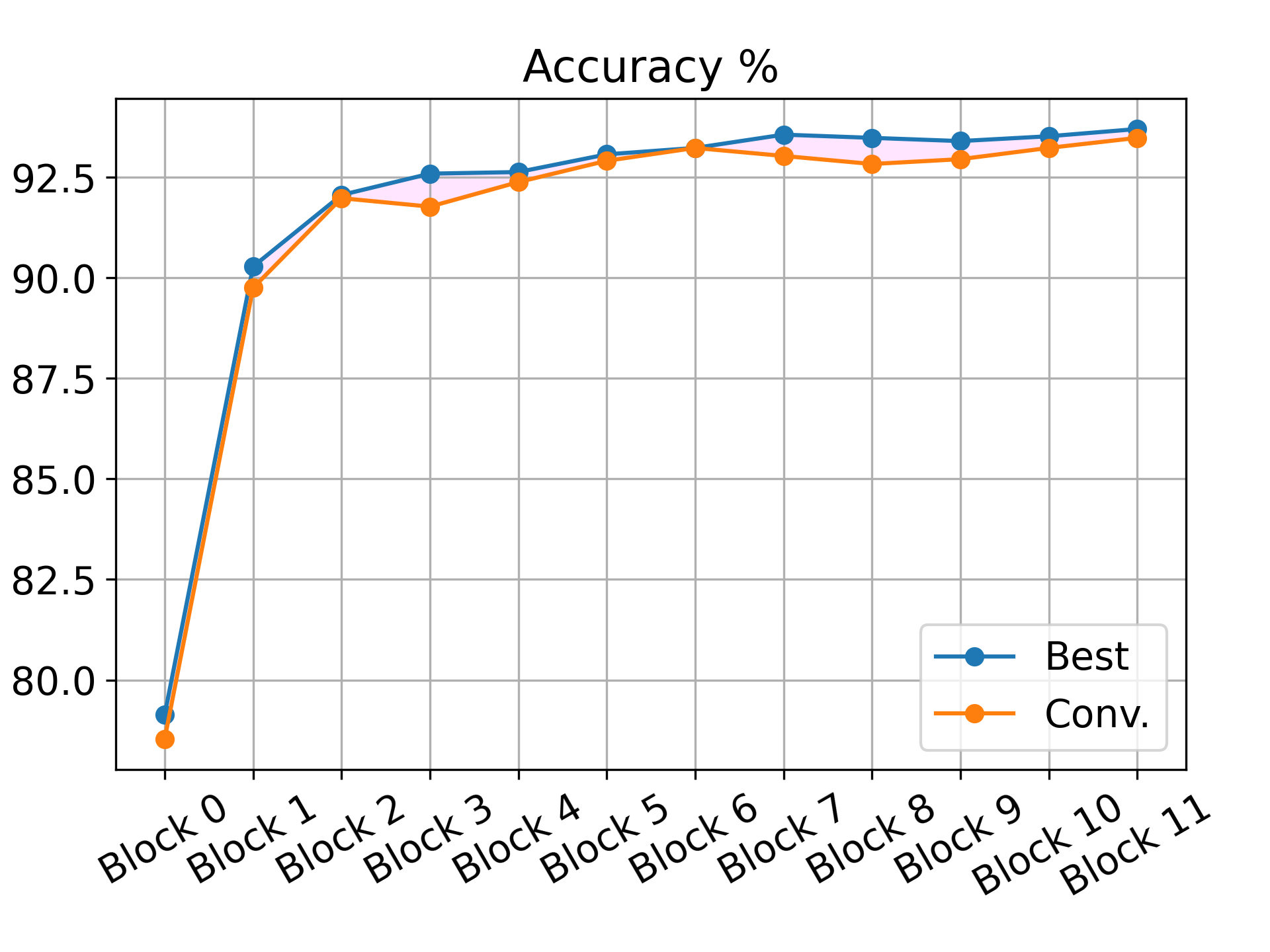}
    \caption{Accuracy graph of truncated copies of the same model. Each entry specifies the layer from which the feature vector is derived (all succeeding layers are discarded for that run). We observe that the model maintains higher accuracy than the baseline, even with a fraction of the original model's parameters.}

    \label{fig:accuracy}
\end{figure}

The classification token in the transformer model progressively accumulates information in order to get a high-level understanding of the input shape. In our second experiment we try to evaluate the representation attained through intermediate layers. In Figure \ref{fig:accuracy} we use the same pretrained backbone, but use only a subset of its transformer blocks. Numeric results are presented in Table \ref{t:accuracy} . When removing up to 5 layers, the model manages to retrain an accuracy above the baseline. Interestingly, that's the same layer in which the transition from location-specific to uniform attention becomes clearer. This indicates that the class cluster separation is possible even in earlier layers, and any additional information simply contributes to refining these clusters.

\begin{table}
\caption{Classification results on ModelNet40 when using the CLS token from previous layers as global descriptor of the 3D shapes. The network is trained with our unfreezing strategy, that is training only the cls head for the first 250 epochs.}
\label{t:accuracy}
\begin{tabular}{l|c c c  }
Model & Point-BERT & Point-MAE & Ours(full depth)  \\
\hline
Acc. & 92.70 & 93.19 & 93.70 \\ 
\hline
\end{tabular}
\begin{tabular}{l|c c c c c }
Without Layers & -5 & -4 & -3 & -2 & -1 \\
\hline
Acc. & 93.23 & 93.56 & 93.48 & 93.40 & 93.52 \\
\end{tabular}
\end{table}

\section{Discussion}

Using our proposed unfreezing strategy, we manage to outperform our baseline model, as well as models that use transformers in a similar way. We do not use voting unless explicitly mentioned for fair comparisons, nor do we include random transforms in the validation set. The method we suggest combined with our learning rate scheduling policy yield stable training with very close convergence points and weaker fluctuations.

Given sufficient training time, task-specific knowledge must be incorporated into the backbone in order to improve performance, but only when the head network has been trained enough for the weights to move in the appropriate direction. The specific intervals can vary depending on the task and dataset. For instance, data with background obtained from scans (in ScanObjectNN) require more train time for the backbone, so the optimal unfreezing point is around 200 epochs instead of 250. We also find that when the data distribution is fairly different from the pretraining data, it is more beneficial to pretrain our network for a few epochs in the new distribution instead of unfreezing the backbone earlier (Table \ref{t:ScanObjectNN}).

In our explainability experiments, we observe that point neighborhoods with sharp geometric features are generally more likely to be attended to by the classification token. This is an indicator of good pretraining, as sharp features are more relevant descriptors of the sample's class.
We also notice that the transition from local to global seems characteristic of transformers. This property is experimentally backed in the image domain \citep{NEURIPS2021_652cf383}, and our own experiments confirm that this holds true in point clouds as well.
An additional surprising finding is the contrast between the behaviors of two finetuned models, one that has been pretrained and one that has not. In the first case the human eye can clearly capture a transition from local to global, whereas in the second case the attention scores are mostly random, containing no identifiable pattern.  

Finally, we remark that doubling the volume of data causes more attention heads to attend locally and introduces accuracy boosts in the classification task, albeit small. In \citep{NEURIPS2021_652cf383} they observe similar behavior but with even greater impact, which is likely attributed to the significantly larger scale of both the datasets and models. Nevertheless, in both cases it suggests that local information aggregation is a property linked with good performance.

\section{Future Work}

 Our analysis in section \ref{sec:6} demonstrates that contrastive learning exhibits useful properties, namely, symmetry and the ability to train a classification token in the pretraining stage. Consequently, we believe that contrastive learning can potentially outperform MAE in several tasks, given enough training time and data. In fact, we theorize that due to their distinct characteristics, some pretraining schemes may offer greater advantages for a specific downstream task than others. In future work, we plan on experimenting on a framework for determining which pretraining scheme to use, in order to utilize our models to their fullest potential. 


\appendices

\section{\break MoCo \& MAE Hybrid Pretraining}
We devise a pretraing sceme that combines both core concepts from MoCo and MAE. The main idea is that the reconstruction loss from MAE will help the backbone converge faster, being a more straightforward task, and the contrastive loss will help in training the CLS token and slightly contribute to the backbone features as well. The scheme is realized through a weighted sum of both losses, appropriately scaled so that they are in the same order of magnitude. 

Specifically, the standard reconstruction loss presented in [33] is applied ($L_{rec}$) and a contrastive loss, $L_{con}$, as in [20] is added, using a queue size of $4096$ encoded samples. The final loss function is realized as:

$$L = L_{rec} + w_c \cdot L_{con}$$

Where $w_c$ was set to $10^{-2}$ to bring the two losses to the same order of magnitude.
We also experimented with a smaller weight($10^{-3}$), as the MoCo loss is responsible mainly for the CLS token. Both approaches, however, produced similar results.  

Although the hybrid model's accuracy reaches an unimpressive $91.8$\% score, we believe that with further research it is possible to create a model that makes the best of both worlds (masked autoencoding and contrastive learning).

\begin{figure}[!t]
    \centering
    \includegraphics[width=3.25in]{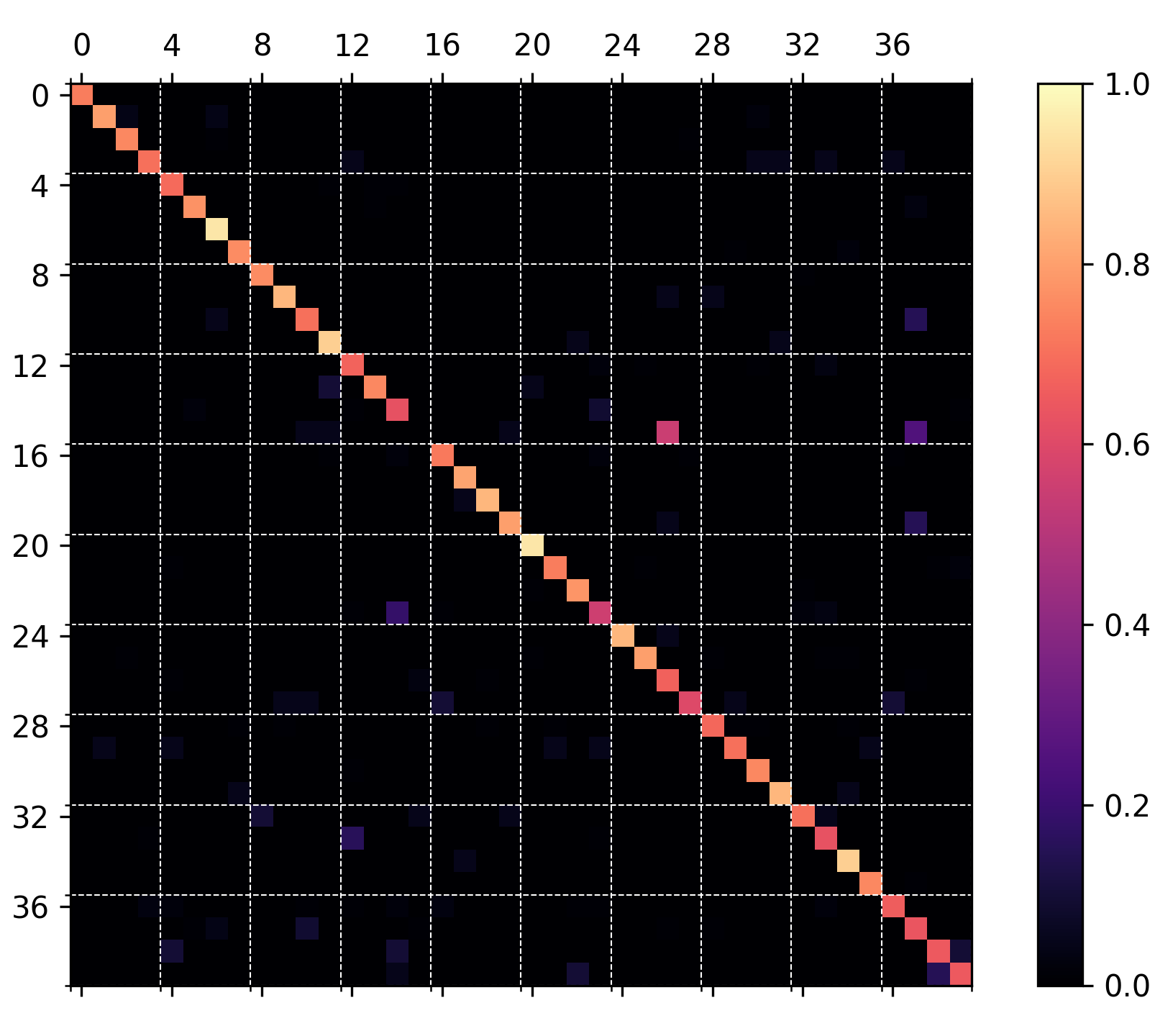}
    \caption{Confusion matrix visualization for the validation set of ModelNet40. Each row has been normalized through division by the number of samples of that class.}
    \label{fig:confusion_matrix}
\end{figure}

\subsection{Strategic unfreezing}

As briefly discussed in the main paper (due to space constraints), the finetuning data from ModelNet40 have already been seen during the pretraining stage and therefore the distribution of said data has been captured by the backbone. Therefore, it is sufficient to only train the backbone for the last 50 epochs, where the learning rate is small enough so that no overblown weight updates happen. 

The ScanObjectNN is probably an even better indicator of this strategy. We observe that in table \ref{t:ScanObjectNN}, when training on objects without background the same unfreezing point yields the best results. On the other hand, when background is included, unfreezing at 200 epochs is optimal and obtains even better accuracy. While the higher accuracy can easily be attributed to the extra information contained in the background itself, it is important to notice that the background changes the data distribution significantly. This change, coupled with the fact that the data has been obtained through scanning, leads us to think that 50 epochs is no longer enough for the backbone to adjust to the new distribution. 

This argument is further verified by finetuning on the hard variant of ScanObjectNN. Sure enough, 100 epochs of training with the unfrozen backbone makes a staggering $+1.18\%$ difference in accuracy compared to 50 epochs, however it is still far from the optimal, which is to \textit{pretrain} the network for a few epochs on this dataset. In our experiments, 40 epochs of pretraining are adequate to reach an accuracy of $85.25\%$, that surpasses our baseline Point-MAE.

Generally speaking, there is no explicit way to determine the exact epoch, at which the unfreezing yields optimal results. However, our proposed "strategic unfreezing" method is a practical way of iteratively improving the model at hand, by providing concrete proof regarding the similarity with the original pretrained version of the model. Consequently, it allows the researcher to make informed design choices and gain an overall deeper understanding. \color{black}

\subsection{Optimization and learning rate policy}
For both our optimizer and learning rate we follow one of [33]'s suggestions. For our finetuning we use the \textit{AdamW} optimizer with a peak learning rate of $\lambda = 5 \cdot 10^{-4}$ and a weight decay of $W_d = 0.05$. We utilize the \textit{cosine annealing learning rate scheduler}, updating during every step for a total training of 300 epochs. A \textit{linear warmup} is also performed, with an initial learning rate $\lambda_i = 10^{-6}$ for a duration of 10 epochs in total. The exact same setup is also used for pretraining, with the exception of peak learning rate, being $\lambda = 10^{-3}$. The batch size is 128 for pretraining and 32 for finetuning. 

\subsection{Data Preprocessing}
All data samples are normalized to be inside the unit sphere. For data augmentations we apply random sampling of the point clouds and anisotropic scale, following the paradigm of our baseline.   

\subsection{Hardware}

All models were trained on a single \textit{NVIDIA RTX 4090} GPU. It should be noted that the memory requirements did not exceed 16GB. The training times were (approximately) 16 hours for pretraining on the combined dataset. For finetuning on our best model (unfreeze 250) the training time was 2 hours on ModelNet40 and 1 hour on ScanObjectNN. Note that when unfreezing at an earlier stage, the total time is lengthened due to the weight updates happening inside the backbone. It takes the model approximately $36$ ms to process a batch with 16 samples.

\section{\break Class Clusters}

\begin{figure}[!t]
    \centering
    \includegraphics[width=3.05in]{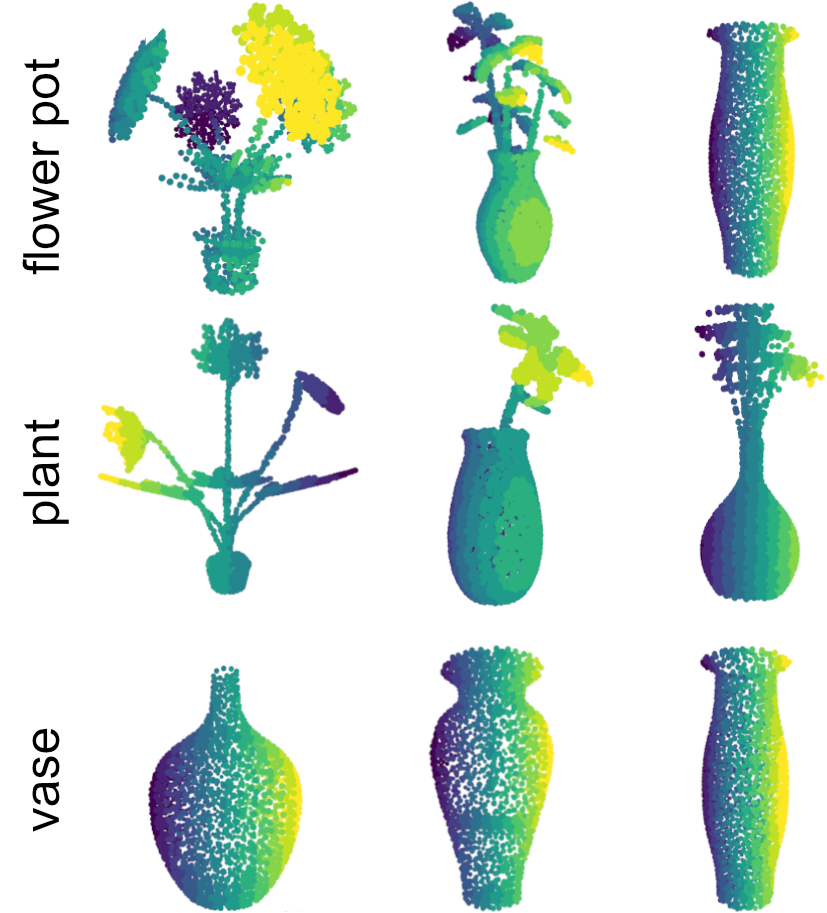}
    \caption{Visualization of samples from problematic classes of ModelNet40. First, second and third rows correspond to \textit{"flower pot", "plant"} and \textit{"vase"} respectively.}
    \label{fig:trash}
\end{figure}

In order to get a better idea of the causes behind misclassified objects we present a confusion matrix (figure \ref{fig:confusion_matrix}) generated from samples of ModelNet40 validation set. The most striking entry is the 15th category, \textit{"flower pot"}, being mistaken for classes 26, \textit{"plant"}, and 37, \textit{"vase"}. These categories contain an unreasonably large amount of overlap, e.g. pots have plants, plants have pots and a lot of pots have a vase-like shape (figure \ref{fig:trash}). Therefore, it is hardly a surprising result that these objects are being misclassified, and this particular error is not a good descriptor of the model's performance.

\begin{figure*}
    \centering
    \includegraphics[scale = 0.10]{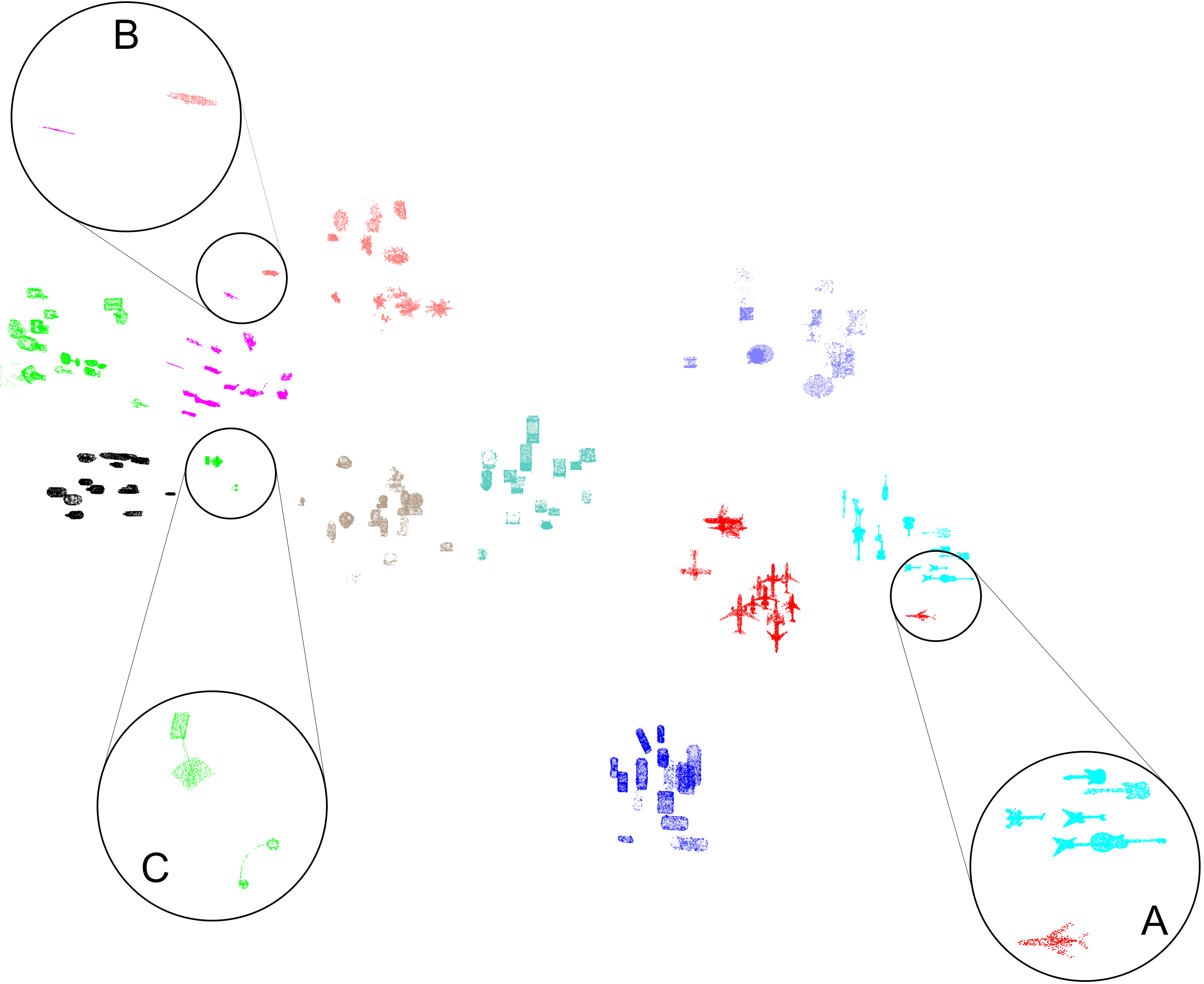}
    \caption{Clustering performed by the model in the feature space for 10 arbitrary classes. The model has been pretrained using the 
    \textbf{MAE} pipeline. T-SNE is used to project the models two the 2-dimensional space. (A) A sample from the \textit{airplane} class is placed near the \textit{guitar} cluster. An "excusable" choice as both share a lengthy shape and sharp protruding parts. (B) Two shape from the \textit{person} and \textit{plant} classes are placed further from their clusters, and near each other, as they have a near non-existent depth. (C) Two uniquely shaped lamps have been assigned to a cluster of their own.}
  \label{fig:tsnemae}    
    \includegraphics[scale = 0.10]{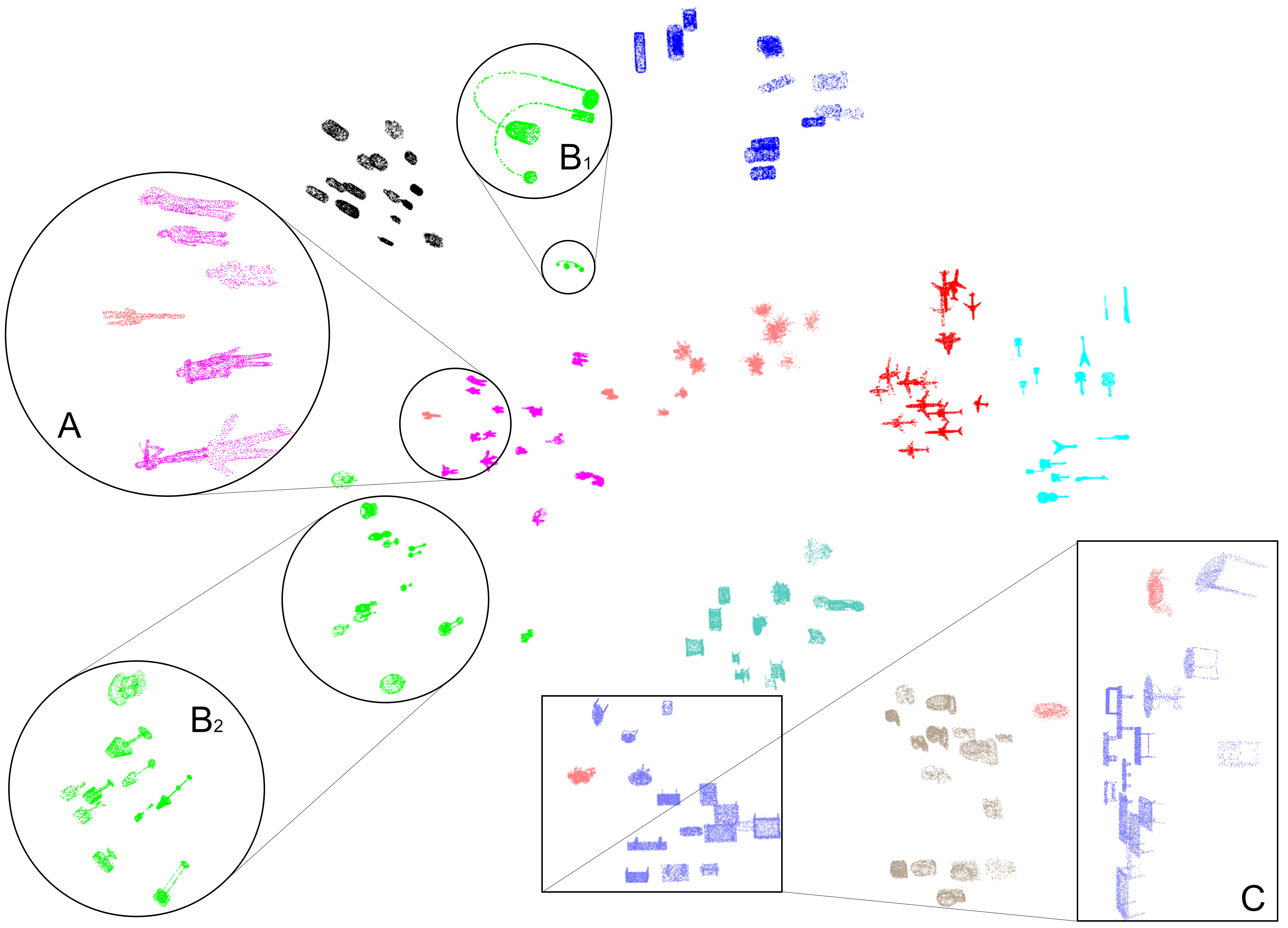}
    \caption{Clustering performed by the model in the feature space for 10 arbitrary classes. The model has been pretrained using the 
    \textbf{MoCo} pipeline. T-SNE is used to project the models two the 2-dimensional space. (A) A tall cactus is clustered together with the \textit{people} class. (B1), (B2), Two lamps with distinct characteristics have been assigned to a separate cluster. (C) A plant with hanging leaves is placed along with the \textit{table} cluster, most likely because the leaves share a similar structure with table legs.  }
  \label{fig:tsnemoco}     
\end{figure*}

In figures \ref{fig:tsnemae} and \ref{fig:tsnemoco} (MAE and MoCo respectively) we visualize the feature vectors of 15 shapes from 10 arbitrary classes randomly sampled from the validation set of ModelNet40 and projected onto the 2d space by using T-SNE. Naturally, the test accuracy is the sole indicator of how well the clustering is performed. However, by using this technique we are hoping to understand whether the clustering/classification operates within the realms of human logic. For this purpose, we are mostly interested in the misclustered samples. 

All in all, these mistakes seem plausible from a human perspective. In particular, the model seems to be "fooled" by objects that have similar height/width or are composed by roughly equivalent parts. This is hardly a fault however, as it only decreases the score on the ModelNet40 benchmark, whilst achieving greater generalization ability. 

\bibliographystyle{IEEEtran}
\bibliography{bibliography}

\begin{biography}[{\includegraphics[width=1in,height=1.25in,clip,keepaspectratio]{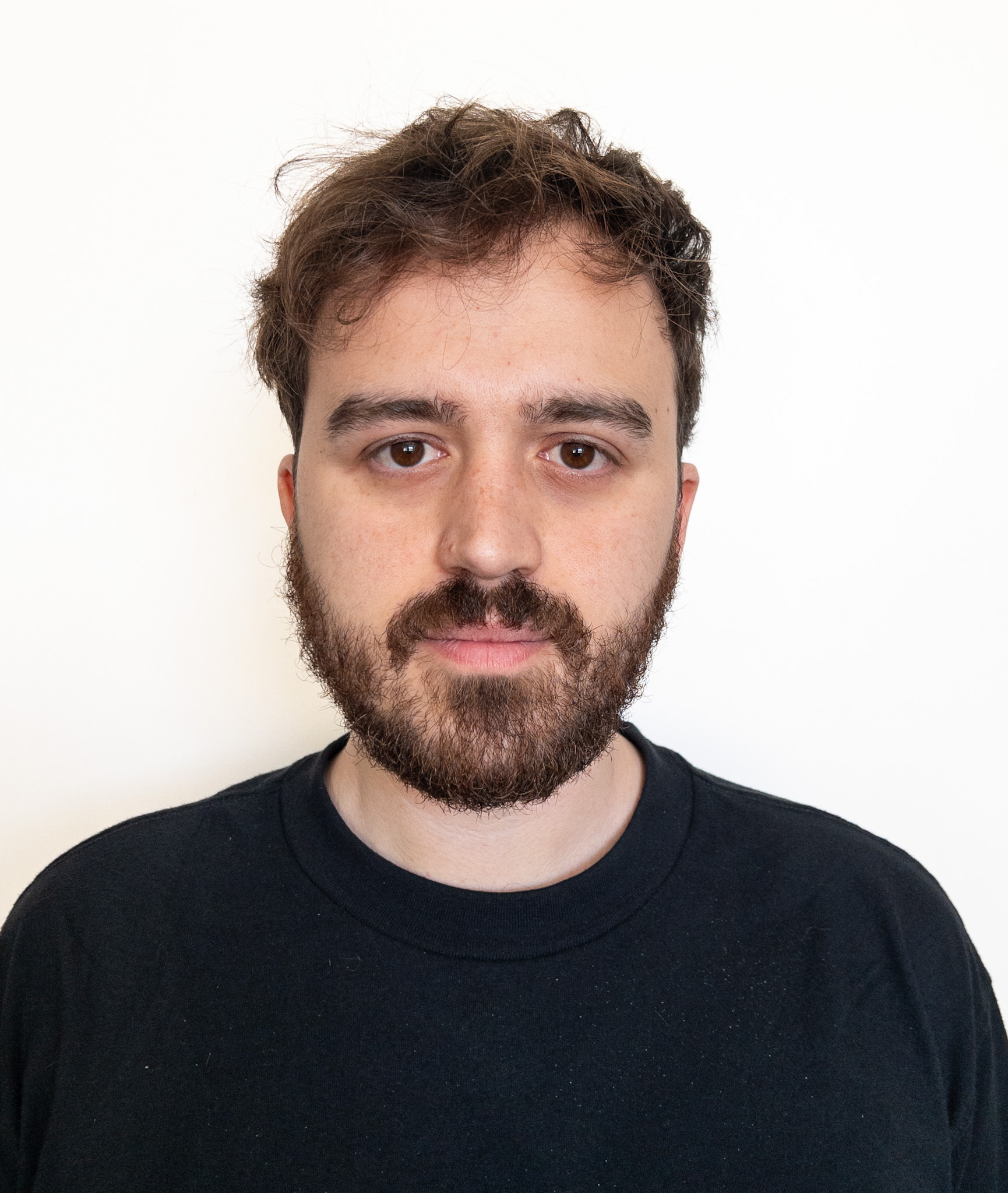}}]{Ioannis Romanelis}
received his Electrical and Computer Engineering diploma in 2021 at the University of Patras. During the same year, he enrolled for a PhD at the same department under the supervision of Professor Konstantinos Moustakas and joined the Visualization and Virtual Reality (VVR) group. His main research interests include computer vision, deep learning, point cloud processing, 3D scene understanding, and explainable AI.
\end{biography}

\begin{biography}[{\includegraphics[width=1in,height=1.25in,clip,keepaspectratio]{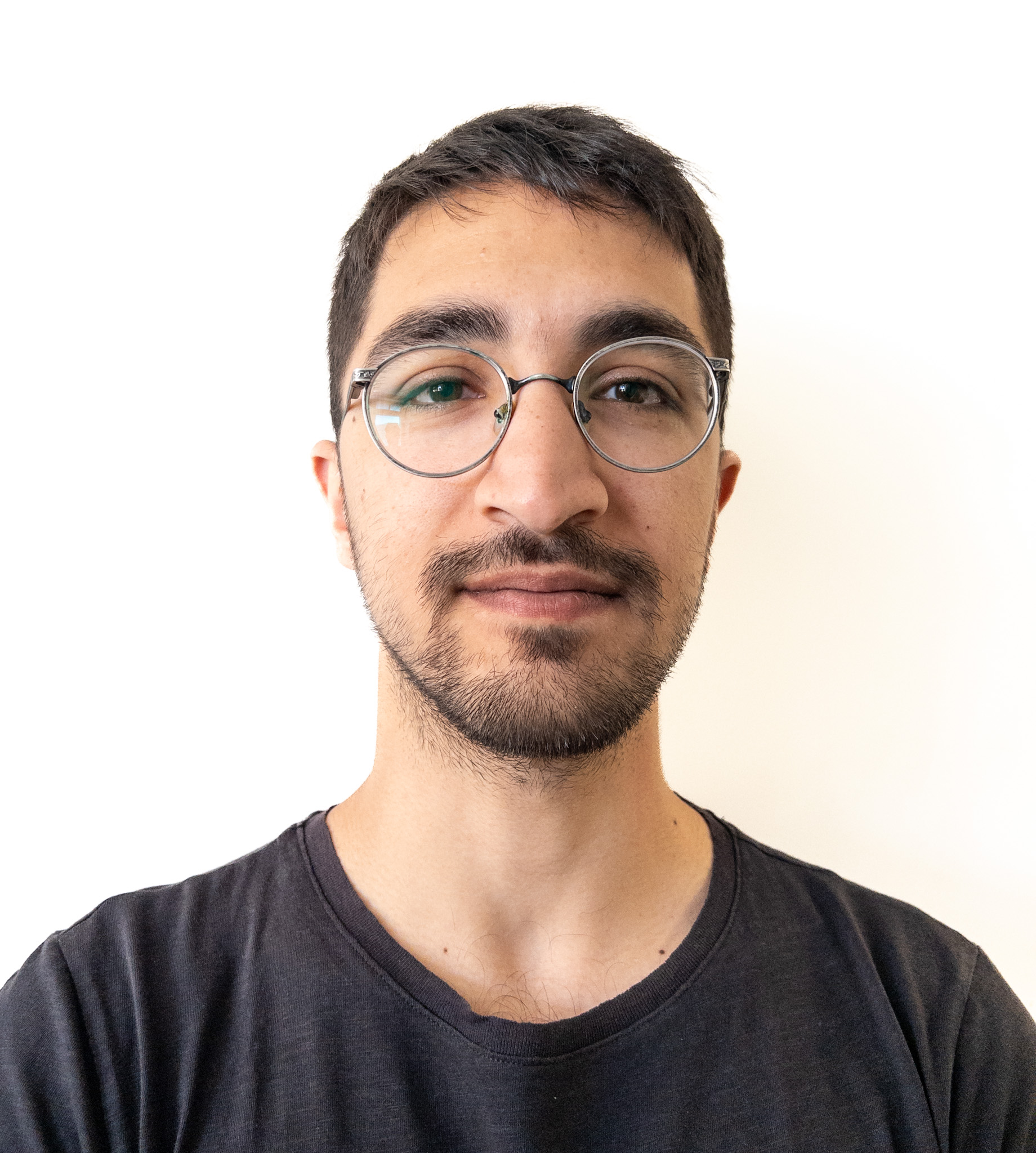}}]{Vlassis Fotis}
received his Electrical Engineering and Computer Technology degree in 2021 at the university of Patras. During the same year he enrolled as a PhD candidate in the same department  under the supervision of Professor Konstantinos Moustakas and joined the Visualization and Virtual Reality (VVR) group. His main research interests include (but are not limited to) computer vision, theoretical deep learning, 3D scene understanding and geometry processing.
\end{biography}

\begin{biography}[{\includegraphics[width=1in,height=1.25in,clip,keepaspectratio]{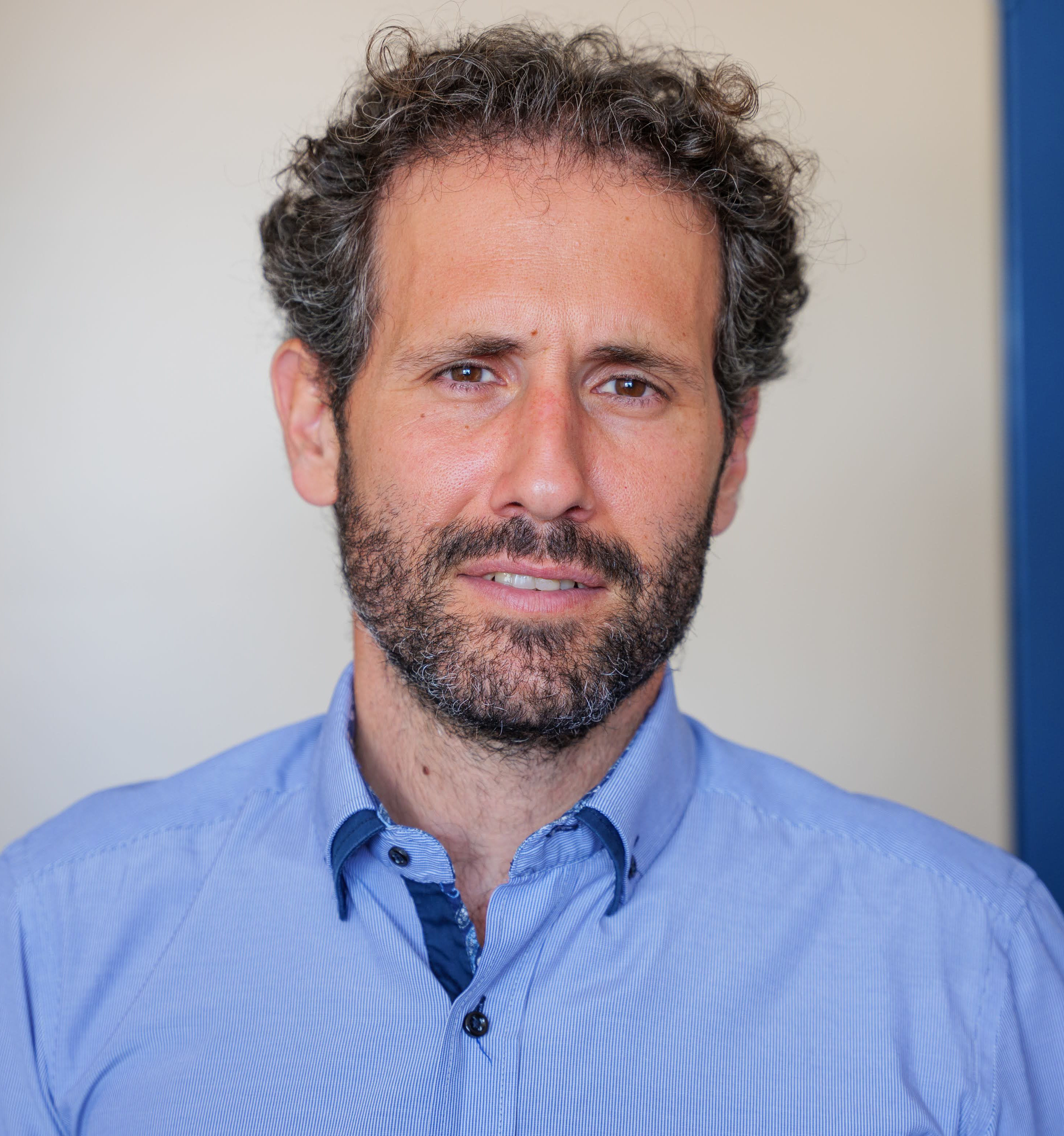}}]{Konstantinos Moustakas} 
Konstantinos Moustakas received the Diploma degree and the PhD in electrical and computer engineering from the Aristotle University of Thessaloniki, Greece, in 2003 and 2007 respectively. During 2007-2011 he served as a post-doctoral research fellow in the Information Technologies Institute, Centre for Research and Technology Hellas. He is currently a Professor at the Electrical and Computer Engineering Department of the University of Patras, Head of the Visualization and Virtual Reality Group, Director of the Wire Communications and Information Technology Laboratory and Director of the MSc Program on Biomedical Engineering of the University of Patras. He serves as an Academic Research Fellow for ISI/Athena research center. His main research interests include virtual, augmented and mixed reality, 3D geometry processing, haptics, virtual physiological human modeling, information visualization, physics-based simulations, computational geometry, computer vision. During the latest years, he has been the (co)author of more than 300 papers in refereed journals, edited books, and international conferences. His research work has received several awards. He has participated in more than 30 research and development projects funded by the EC and the Greek Secretariat of Research and Technology, while he has served as the coordinator or scientific coordinator in 4 of them. He has also been a member of the organizing committee of several international conferences. He is a senior member of the IEEE, the IEEE Computer Society and member of Eurographics.
\end{biography}

\begin{biography}[{\includegraphics[width=1in,height=1.25in,clip,keepaspectratio]{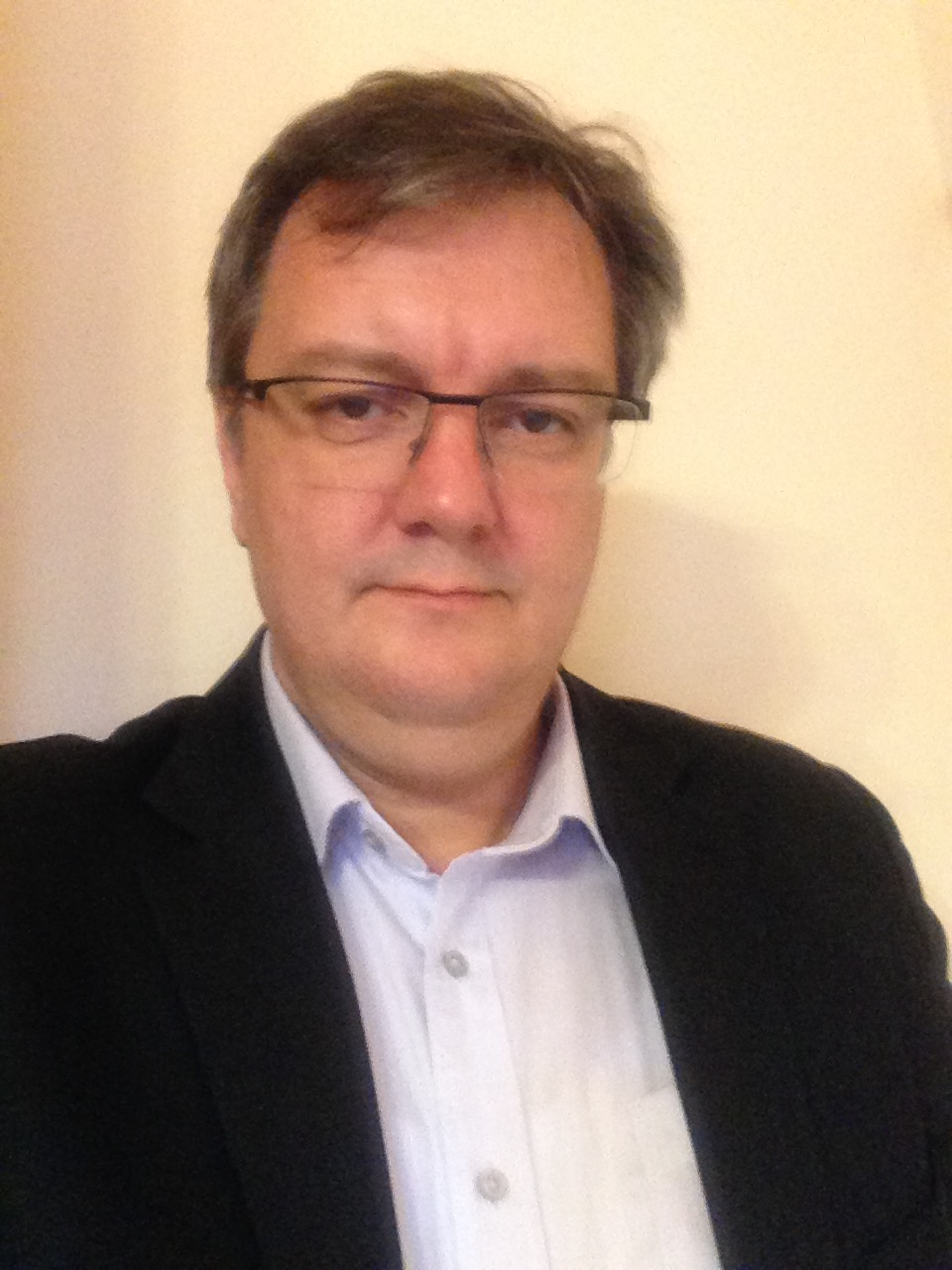}}]{Adrian Munteanu}
Adrian Munteanu (M’07) is professor at the Electronics and Informatics (ETRO) department of the Vrije Universiteit Brussel (VUB), Belgium. He received the MSc degree in Electronics and Telecommunications from Politehnica University of Bucharest, Romania, in 1994, the MSc degree in Biomedical Engineering from University of Patras, Greece, in 1996, and the Doctorate degree in Applied Sciences (Maxima Cum Laudae) from Vrije Universiteit Brussel, Belgium, in 2003.

From 2004 to 2010, he was post-doctoral fellow with the Fund for Scientific Research – Flanders (FWO), Belgium, and since 2007, he is professor at VUB. His research interests include image, video and 3D graphics compression, 3D video, deep-learning, distributed visual processing, error-resilient coding, and multimedia transmission over networks. 

Adrian Munteanu is the author of more than 400 journal and conference publications, book chapters, and contributions to standards and holds seven patents in image and video coding. He is the recipient of the 2004 BARCO-FWO prize for his PhD work, a (co-)recipient of the Most Cited Paper Award from Elsevier for 2007, and of ten other scientific prizes and awards. Adrian Munteanu served as Associate Editor for IEEE Transactions on Multimedia and currently serves as Associate Editor for IEEE Transactions on Image Processing.
\end{biography}

\EOD

\end{document}